\documentclass[10pt,journal,compsoc]{IEEEtran}
\ifCLASSOPTIONcompsoc
\else
\fi

\usepackage[OT1]{fontenc}
\usepackage{lipsum}
\usepackage{url}
\usepackage{amsmath}
\usepackage{bbm}
\usepackage{booktabs}
\usepackage{graphicx}
\usepackage{color}
\usepackage{float}
\usepackage{array}
\usepackage{amssymb}
\usepackage{multirow,makecell}
\usepackage[implicit=false]{hyperref}
\usepackage{ragged2e}

\begin{document}
	\title{Open World Entity Segmentation}
	\author{Lu Qi,
			Jason Kuen$^\dagger$,
			Yi Wang,
			Jiuxiang Gu,
			Hengshuang Zhao,
			Philip Torr,
			Zhe Lin,
			Jiaya Jia
\IEEEcompsocitemizethanks{
\IEEEcompsocthanksitem L~.Qi and J.~Jia are with the Department of CSE, The Chinese University of Hong Kong. Jason Kuen, J~.Gu and Z~.Lin are with the Adobe Research. Y~.Wang is with the Shanghai AI Lab. H.~Zhao is with the Department of Computer Science at The University of Hong Kong. Philip Torr is with the Department of Engineering Science, University of Oxford. 
\IEEEcompsocthanksitem Jason Kuen$^\dagger$ is the corresponding author. The contact emails are qqlu1992@gmail.com (Lu Qi) and kuen@adobe.com (Jason Kuen).
}
}

	\definecolor{mypink3}{cmyk}{0, 0.7808, 0.4429, 0.1412}
	% \newcommand{\lq}[1]{{\color{mypink3}{\bf\sf [lq: #1]}}}
	
	% The paper headers
	% \markboth{Journal of \LaTeX\ Class Files,~Vol.~14, No.~8, August~2015}%
	% {Shell \MakeLowercase{\textit{et al.}}: Bare Demo of IEEEtran.cls for Computer Society Journals}

\IEEEtitleabstractindextext{%
\begin{abstract} \justifying
We introduce a new image segmentation task, called Entity Segmentation (ES), which aims to segment all visual entities (objects and stuffs) in an image without predicting their semantic labels. By removing the need of class label prediction, the models trained for such task can focus more on improving segmentation quality. It has many practical applications such as image manipulation and editing where the quality of segmentation masks is crucial but class labels are less important. We conduct the first-ever study to investigate the feasibility of convolutional center-based representation to segment \textit{things} and \textit{stuffs} in a unified manner, and show that such representation fits exceptionally well in the context of ES. More specifically, we propose a CondInst-like fully-convolutional architecture with two novel modules specifically designed to exploit the class-agnostic and non-overlapping requirements of ES. Experiments show that the models designed and trained for ES significantly outperforms popular class-specific panoptic segmentation models in terms of segmentation quality. Moreover, an ES model can be easily trained on a combination of multiple datasets without the need to resolve label conflicts in dataset merging, and the model trained for ES on one or more datasets can generalize very well to other test datasets of unseen domains. The code has been released at \href{https://github.com/dvlab-research/Entity}{https://github.com/dvlab-research/Entity}.
\end{abstract}
		
\begin{IEEEkeywords}
Image Segmentation, Class-Agnostic, Open-World, Cross-Dataset.
\end{IEEEkeywords}}
	
% make the title area
\maketitle

\IEEEdisplaynontitleabstractindextext
\IEEEpeerreviewmaketitle
\IEEEraisesectionheading{\section{Introduction}}
In recent years, image segmentation tasks (semantic segmentation~\cite{long2015fully,zhao2017pyramid,chen2017rethinking,chen2018encoder,zhao2018psanet,huang2019ccnet,liu2019auto}, instance segmentation~\cite{dai2016instance,li2016fully,he2017mask,liu2018path,xie2020polarmask,tian2020conditional,wang2019solo,wang2020solov2,qi2020pointins}, and panoptic segmentation~\cite{kirillov2019panoptic,kirillov2019panopticfpn,xiong2019upsnet,li2020fully,carion2020end}, \textit{etc.}) have received great attention due to their diverse applications~\cite{shu2014human,morrison2018cartman,qi2019amodal,rong2020frankmocap,cao2018pose,rong2019delving} and strong progress made possible by deep learning~\cite{krizhevsky2017imagenet,simonyan2014very,szegedy2016rethinking,szegedy2016inception,he2016res,huang2017densely,hu2018squeeze}. Most image segmentation tasks share the common goal of automatically assigning each image pixel to one of the predefined semantic classes. One of the key application areas of image segmentation is image manipulation~\cite{sun2005image,wang2018inpainting,isola2017image,wang2017high,yu2018generative,yu2019free} and editing~\cite{barnes2009patchmatch,darabi2012image,levin2004colorization,perez2003poisson}. Segmentation techniques have revolutionized image manipulation and editing applications by enabling users to operate directly on semantically-meaningful regions, as opposed to primitive image elements such as pixels and superpixels.

\begin{figure*}[t!]
\begin{center}
\includegraphics[width=\linewidth]{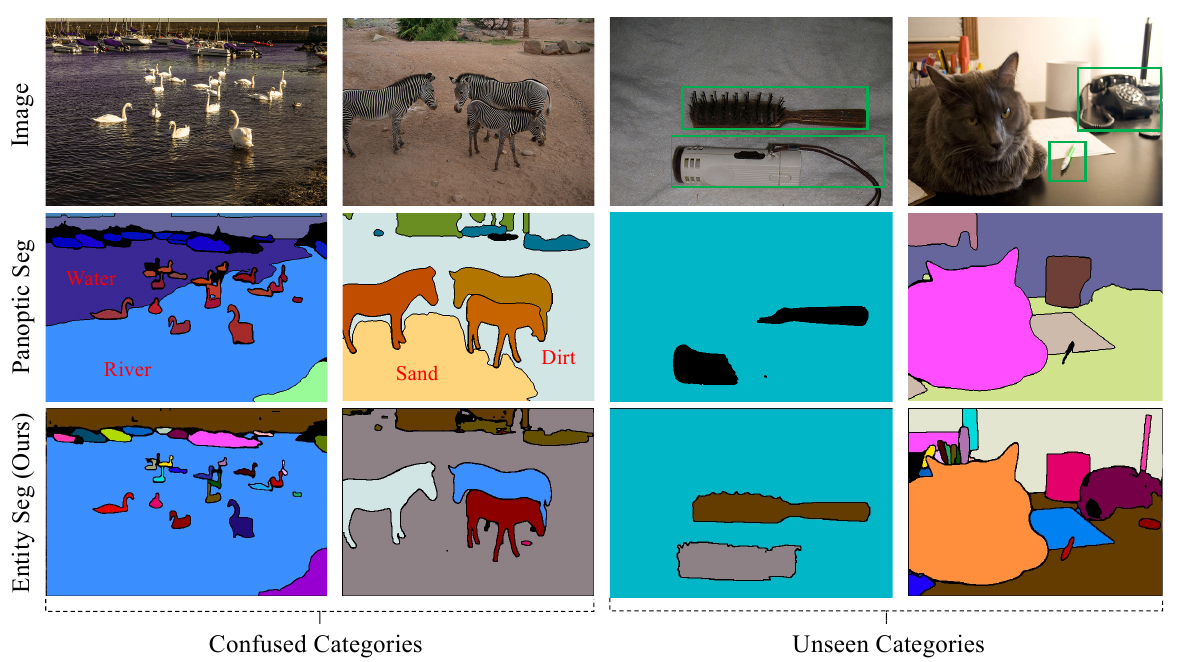}
\end{center}
\caption{The first two columns show the examples of class confusion. Even with a state-of-the-art panoptic segmentation network\protect\footnotemark[1], there tends to be two or more masks with semantically-overlapped classes ({\em e.g.,} {\em river} \& {\em water}, {\em sand} \& {\em dirt}) predicted for a single entity. The last two second columns illustrate that the network has troubles segmenting unseen objects like {\em afro pick}, {\em pencil}, {\em telephone}.}
\label{fig:motivation}
\end{figure*}

Although image segmentation holds much promise for user-friendly image manipulation and editing, there are two major weaknesses in image segmentation that adversely affect the image manipulation/editing experience: 1) class label confusion (often caused by having many predefined class labels); 2) lack of generalization for unseen classes. In Fig.~\ref{fig:motivation}, we present the segmentation results from a state-of-the-art panoptic segmentation method \cite{li2020fully} that demonstrate the two weaknesses separately. These weaknesses can be largely attributed to the standard practice of training models to segment strictly based on the predefined classes.

In image manipulation and editing applications, where class labels are typically not required, the conventional class-specific image segmentation may be sub-optimal and could introduce unnecessary class-related issues. Is there a better alternative to class-specific image segmentation? Yes, it comes from omitting the classification subtask and focusing solely on the segmentation subtask, in a similar spirit to the class-agnostic RPN~\cite{ren2015faster}. This is analogous to infants capable of distinguishing objects by shapes and appearances without knowing the object names~\cite{infant}. Without assigning class labels, the class confusion issue can be largely alleviated. Besides, as demonstrated in object detection~\cite{jaiswal2021class, rahman2018zero}, class agnosticism provides a strong generalization to unseen classes. Furthermore, 
it is likely to relieve the model's burden of trading off segmentation mask quality against classification performance.

To this end, we propose a new image segmentation task named Entity Segmentation (ES) which aims to generate
class-agnostic segmentation masks of an image. We leverage the existing panoptic segmentation datasets~\cite{lin2014microsoft,zhou2017scene,cordts2016cityscapes} with both instance (\textit{thing}) and non-instance (\textit{stuff}) masks, but universally treat any of the annotated \textit{thing} and \textit{stuff} masks as a class-agnostic \textit{\textbf{entity}}. Here, we are noting that entity segmentation is advanced version of panoptic segmentation. As introduced in Marr's Vision Book~\cite{Marr1982Vision}, the human vision system is internally class-agnostic, which could recognize entity without the sense of understanding its use and purpose. Since the popular class-specific Panoptic Quality (PQ) metric~\cite{kirillov2019panoptic} is not directly applicable to our task, we introduce a class-agnostic metric AP$_\text{e}$ with a strict non-overlapping mask constraint. This constraint aligns well with image editing/manipulation applications that expect each image pixel only correspond to a single entity.

Given our proposed new task and evaluation metric, we aim to devise a conceptually simple and effective framework to tackle the task. The center-based output representation (\textit{i.e.,} representing objects by their centers) has enabled recent one-stage object detectors~\cite{tian2019fcos,lin2017focal} to achieve strong performance and fast training. Inspired by that, we carried out the first-ever study to investigate the feasibility of using such a representation uniformly for all entities including \textit{stuffs}. Our study reveals a strong and previously-unknown evidence that a fully-convolutional network paired with the center-based representation can effectively handle the task. We are the first to challenge the common understanding that the difficult-to-train Transformer decoder \cite{carion2020end,cheng2021per} is necessary to unify the representation for {\em things} and {\em stuffs}.

Based on our findings, we introduce a fully-convolutional segmentation framework inspired by CondInst \cite{tian2020conditional} to detect and segment all entities (regardless of \textit{thing} or \textit{stuff}) in a unified manner. To better exploit the unique \textit{class-agnostic} and \textit{non-overlapping} requirements of ES, we propose a global kernel bank module and an overlap suppression module that substantially improve the framework and thus boost the performance of the ES task. The global kernel bank module generates mask kernels to exploit some common properties (\textit{e.g.,} textures, edges) shared by many entities, while the overlap suppression module encourages the predicted masks not to overlap with each other.

\footnotetext[1]{PanopticFCN~\cite{li2020fully} with ResNet101~\cite{he2016res} backbone and Deformable Convolution v2~\cite{dai2017deformable,zhu2019deformable}.}

Extensive experiments on the challenging COCO dataset~\cite{lin2014microsoft} show the effectiveness of our proposed method. Furthermore, we study the generalization ability of our COCO-trained model through a cross-dataset evaluation on ADE20K~\cite{zhou2017scene}.
Our model shows superior performance both quantitatively and qualitatively on such a dataset despite that it has not been trained on it. As shown in Fig.~\ref{fig:generalization}, the model is able to correctly segment the entities belonging to unseen classes. These suggest that the models trained for the proposed ES task are naturally capable of open-world image segmentation.

Overall, our contributions are summarized as follows:
\begin{itemize}
\item We propose a task called Entity Segmentation which aims to segment every visual entity without predicting its class label or identifying it as \textit{thing} or \textit{stuff}. 
\item We conduct the first-ever study to investigate the feasibility of convolutional center-based representation for various tasks and surprisingly find that all entities ({regardless of} {\em thing} or {\em stuff}) can be effectively represented by such a representation in a unified manner.
\item Our findings lead us to devise a fully-convolutional segmentation framework inspired by CondInst. We introduce two novel modules to our framework to better exploit the unique requirements of the ES task.
\item We conduct extensive experiments and demonstrate the remarkable effectiveness and generalization of our proposed task and segmentation framework.
\end{itemize}

\section{Related Work}
\begin{figure*}[t!]
	\begin{center}
		\includegraphics[width=\linewidth]{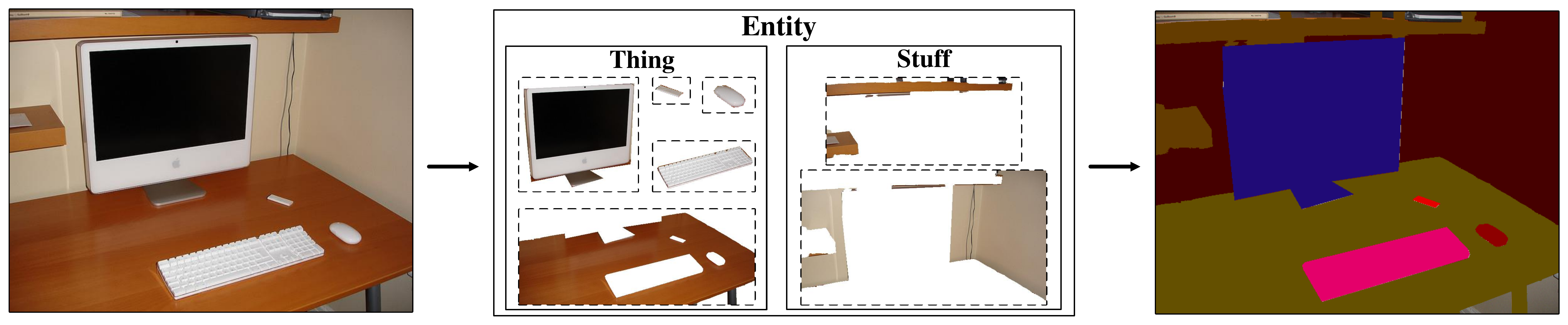} 
	\end{center}
	\caption{The ground-truth annotations of entity segmentation. To convert the annotations of panoptic segmentation to entity segmentation format, we regard each {\em thing} or {\em stuff} as an independent entity, even though some of them may have multiple disconnected parts.} 
	\label{fig:task_definition}
\end{figure*}

\textbf{Image segmentation.} 
Image segmentation is the process of partitioning a digital image into multiple segments (sets of pixels, also known as image objects)~\cite{szeliski2010computer}. The main tasks of image segmentation include salient object detection~\cite{borji2015salient,borji2019salient,jiang2013salient,qin2019basnet,li2016deep}, semantic segmentation~\cite{long2015fully,zhao2017pyramid,chen2017rethinking,chen2018encoder,zhao2018psanet,huang2019ccnet,liu2019auto,yu2018bisenet,yu2018learning,yu2020context,chen2020bi,xing2020malleable}, instance segmentation~\cite{dai2016instance,li2016fully,he2017mask,liu2018path,xie2020polarmask,tian2020conditional,wang2019solo,wang2020solov2,qi2020pointins}, and panoptic segmentation~\cite{kirillov2019panoptic,kirillov2019panopticfpn,xiong2019upsnet,li2020fully,carion2020end,liu2019end}. Salient object detection mimics the behavior of human in focusing on the most salient region/object in an image while ignoring the class it belongs to~\cite{ullah2020brief}. In contrast, semantic/instance/panoptic segmentation aims at densely assigning each image pixel to the one of classes predefined in the training datasets. The segmentation models trained for these tasks are required to make a trade-off between the mask prediction and classification subtasks in terms of performance. Moreover, such tasks tightly couple segmentation and classification. As a result, it is not easy to independently evaluate the segmentation and classification strengths of their segmentation models.

In this work, we provide a new perspective on image segmentation by introducing the entity segmentation task that handles dense image segmentation similarly as semantic/instance/panoptic segmentation, but without the classification aspect akin to salient object detection. This task focuses only on class-agnostic segmentation. It treats every segment (whether {\em thing} or {\em stuff}) as a visual entity. Compared to the existing segmentation tasks, it is more useful for user-friendly image manipulation and editing applications in which the segmentation mask quality is of utmost importance.

\textbf{Object detection.}
Object detection~\cite{girshick2014rich,dollar2015fast,ren2015faster,lin2017feature,lin2017focal,tian2019fcos,dai2016rfcn,qi2021multi} requires detecting objects with bounding box representation. The methods can be partitioned into two types: (1) one-stage~\cite{lin2017focal,tian2019fcos} detectors that detect objects using pixelwise features at dense pixel locations; (2) two-stage detectors~\cite{girshick2014rich,dollar2015fast,ren2015faster,lin2017feature} that first predict class-agnostic region proposals and subsequently perform classification based on region-of-interest (RoI) features and a separate classification head.

In this paper, our method is built upon FCOS~\cite{tian2019fcos}, a standard one-stage detector, to detect entities. Different from the standard definition of {\em object}, entity is a more general concept that includes both objects/{\em things} and {\em stuffs}. To our best knowledge, our method is the first to 
demonstrate that center-based detection can effectively handle all entities in a uniform manner, contrary to the existing practice of separately. In contrast to DETR~\cite{carion2020end} and MaskFormer~\cite{xie2021segformer}, our segmentation framework neither requires an excessively long training duration (12 \textit{vs} 300 epochs) nor strong data augmentation.

\textbf{Open-world detection \& segmentation.}
The open-world setting has been explored in the contexts of object detection and segmentation~\cite{wang2021unidentified,bendale2015towards,liu2019large,DeepMask15,hu2018learning,jaiswal2021class,AR2020theoverlooked,joseph2021open} with the general goal of identifying new object classes given a closed-world training dataset. Generally, previous open-world methods explicitly differentiate {\em unknown} and {\em known} objects by spotting outliers in the embedding space. On the contrary, our entity segmentation task is class-agnostic and does not need to distinguish between {\em unknown} and {\em known} classes, allowing the segmentation models to focus entirely on the segmentation task itself.

\section{Entity Segmentation}
~\label{sec:concept}
\begin{table*}[t]
    \centering
    \small
    \caption{Study on the feasibility of convolutional center-based representation \cite{tian2019fcos,wang2019solo} on different class-agnostic/specific detection tasks that involve \textit{thing} and \textit{stuff}. \textit{Subset} refers to the subset of COCO training/validation set with the annotations from one of: \textit{thing}, \textit{stuff}, or \textit{thing\&stuff} (Entity).
    $\circ$/$\checkmark$ signifies that we do/do-not use class information in the training or evaluation stage. AP$^b$ is the box-based AP.}
    \label{Tab:find}
    \begin{tabular}{c|c|c|ccc|cccl}
        %\toprule
        \cline{1-9}
        \multirow{2}*{Subset} & \multicolumn{2}{c|}{Class-agnostic} &  \multirow{2}*{AP$^{b}$} & \multirow{2}*{{AP}$_{50}^{b}$} & \multirow{2}*{{AP}$_{75}^{b}$} & \multirow{2}*{{AP}$_{S}^{b}$} & \multirow{2}*{{AP}$_{M}^{b}$} & \multirow{2}*{{AP}$_{L}^{b}$} & \\ \cline{2-3}
        & {\em training} & {\em evaluation} & & & & & & \\ \cline{1-9}
        \multirow{3}*{Thing} 
        & $\circ$ & $\circ$ & 37.4 & 56.0 & 39.9 & 15.5 & 36.5 & 50.5 & $\leftarrow$ \texttt{ref} \\
        & $\circ$ & \checkmark & 40.9 & 63.9 & 43.4 & 18.6 & 42.1 & 60.7 \\
        & \checkmark & \checkmark & 41.6 & 64.6 & 43.9 & 18.5 & 42.7 & 61.7 \\ \cline{1-9}%\midrule
        
        \multirow{3}*{Stuff} 
        & $\circ$ & $\circ$ & 23.2 & 35.4 & 23.5 & 1.4 & 5.9 & 26.9 \\
        & $\circ$ & \checkmark & 38.7 & 58.4 & 39.3 & 1.8 & 7.2 & 46.6 \\
        & \checkmark & \checkmark & 39.4 & 60.5 & 39.3 & 1.4 & 6.4 & 47.2 \\ \cline{1-9}%\midrule
        
        \multirowcell{3}{Thing $\&$ Stuff\\ (Entity)} 
        & $\circ$ & $\circ$ & 29.5 & 45.2 & 31.0 & 9.2 & 23.6 & 38.0 \\
        & $\circ$ & \checkmark & 37.6 & 60.2 & 39.3 & 16.5 & 35.0 & 49.5 \\
        & \checkmark & \checkmark & 39.2 & 62.5 & 40.4 & 16.6 & 35.5 & 51.1 \\ \cline{1-9}%\bottomrule
        
    \end{tabular}
\end{table*}

\textbf{Task definition.} The task of entity segmentation (ES) is defined as segmenting all visual entities within an image in a class-agnostic manner. Here, %the
``entity'' refers to either a \textit{thing} (instance) mask or a stuff mask in the common context. This definition is related to the standard and well-accepted definition of “object” which is based on certain objectness properties introduced by the seminal work~\cite{alexe2010object}: (a) a well-defined closed boundary in space; (b) a different appearance from their surroundings. As illustrated in Fig.~\ref{fig:task_definition}, 
an entity can be any semantically meaningful and coherent region in the open-world setting, {\em e.g.,} person, television, wall.  Given the subjective nature of this task, we conduct a comprehensive user study to validate the personal intuitions for the concept of entity in section~\ref{subsec:entity}.

\textbf{Task format.} Given an input image $\mathbf{I} \in \mathbb{R}^{H \times W \times 3}$, the task expects a pixel-wise prediction map $\mathbf{P} \in \{1,...,N\}^{H \times W}$ and a list of confidence scores $\mathbf{S} \in \mathbb{R}^{N}$ as the output that contains the non-overlapping IDs of the predicted $N$ entities and their confidence scores. Ground truth annotations are encoded in an identical manner as the prediction map. There are no semantic class labels involved and all entities are treated equally without the distinction of \textit{thing} and \textit{stuff}.

\textbf{Relationship to similar tasks.} The newly proposed task is focused on the concept of ``entity''%itself
. It is related to but different from several previous tasks. Different from {\em semantic segmentation}, ES is instance-aware. In contrast to {\em instance segmentation}, ES includes \textit{stuff} masks in addition to instance masks. Moreover, unlike the above-mentioned segmentation tasks and the more recent {\em panoptic segmentation}, ES completely omits the class labels and classification functionality.

\textbf{Annotation transformation.} Given the commonalities with panoptic segmentation, the annotations in existing panoptic segmentation datasets can be directly transformed into the format defined for ES. As shown in Fig.~\ref{fig:task_definition},
each \textit{thing} or \textit{stuff} mask is simply regarded as an independent entity. 

\textbf{Evaluation metric.} Due to the non-overlapping requirement in downstream image manipulation and editing applications, we propose a new mask-based mean average precision for measurement, denoted as AP$_{\text{e}}^\text{m}$. AP$_{\text{e}}^\text{m}$ follows closely the AP$^m$ used in instance segmentation~\cite{dai2016instance,li2016fully,he2017mask,liu2018path,xie2020polarmask,tian2020conditional,wang2019solo,wang2020solov2,qi2020pointins}, except that the AP$_{\text{e}}^\text{m}$ gives zero tolerance to mask overlaps of different entities. This simple constraint leads to a lower evaluation number than instance segmentation's AP$^m$, since it makes the metric significantly more sensitive to the class-agnostic duplicate removal~\cite{qi2018sequential} performance and mask quality. In our experiment
, the number drops by 4.7 if AP$_{\text{e}}^\text{m}$ is used in place of AP$^m$.
    
\section{Unified Entity Representation}
\label{sec:analysis}
\begin{figure*}[t!]
\begin{center}
\includegraphics[width=\linewidth]{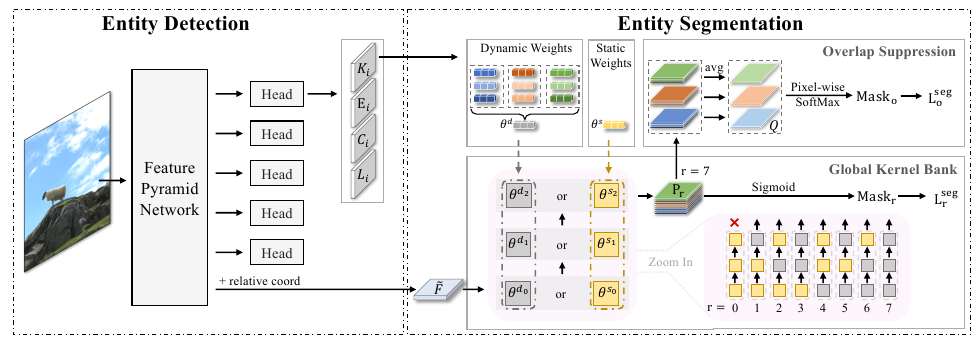}
\end{center}
\caption{The entity segmentation framework. For the part of entity detection, we directly follow the design of FCOS detector~\cite{tian2019fcos}. The entity masks are generated by convolving the low-level feature map $\mathbf{\tilde{F}}$ with dynamic weights $\theta^{d}$. In our method, we propose the global kernel bank and overlap suppression modules to %enhance
exploit class-agnostic and non-overlapping requirements of $\theta^{d}$. The L$^{seg}_o$ here stands for L$^{seg}_{o,r=7}$ which indicates that the global kernel bank's path (r=7) with all dynamic weights is used for the overlap suppresion module.}
\label{fig:framework}
\end{figure*}

The output representation is a defining aspect of any segmentation tasks. Given the close relationship of ES with panoptic segmentation, we first discuss panoptic representation. Due to the different natures of \textit{thing} (instance) and \textit{stuff} (non-instance) classes, existing panoptic segmentation methods usually adopt separate strategies~\cite{kirillov2019panopticfpn,xiong2019upsnet} for the two outputs. In particular, they are handled through a divide-and-conquer pipeline, {\em i.e.,} using a branch that emphasizes explicit localization cues for things, while using another branch that focuses on semantic consistency for \textit{stuff}~\cite{li2020fully}. However, due to the lack of both class labels and \textit{thing}-\textit{stuff} separation in our ES task, the existing convolution-based panoptic segmentation approaches~\cite{kirillov2019panoptic,kirillov2019panopticfpn,xiong2019upsnet,li2020fully,carion2020end,liu2019end,li2021fully} require some form of modifications in order to be applicable to ES. For example, convolution-based methods like PanopticFPN~\cite{kirillov2019panopticfpn}, PanopticFCN~\cite{li2020fully}, and KNet~\cite{zhang2021k} usually adopt separate strategies for thing and stuff segmentation.
When such methods are modified to handle all entities with a single strategy, they tend to lead to a significantly worse accuracy on ES, compared to how well their original designs perform on panoptic segmentation. Even though some transformer-based methods like PanopticDETR~\cite{carion2020end} and MaskFormer~\cite{cheng2021per} are directly capable of building a unified representation for entities, but they are more computationally expensive during training, due to their heavy dependence on stronger data augmentation and longer training time.

As with panoptic segmentation, the question of ``how to represent entities?" is crucial to ES. The center-based representation used by convolutional networks to represent objects by their center locations is a promising form of representation already been proven effective and efficient for object/\textit{thing}-level tasks
({\em e.g.,}object detection~\cite{lin2017focal,tian2019fcos} and instance segmentation~\cite{tian2020conditional,xie2020polarmask,qi2020pointins}), but its feasibility on \textit{stuffs} (and entities) has not been investigated and remains unknown. To answer the aforementioned question, we present the first-ever study\footnote{For the models evaluated in the study, we adopt FCOS~\cite{tian2019fcos} with ResNet50 backbone and default training hyperparameters.} to investigate the feasibility of using center-based representation to represent \textit{stuffs} and entities. The study is carried out through the proxy tasks of class-specific and -agnostic box-level detection. Class-agnostic detection can be seen as the box-level counterpart of ES, and thus the conclusion derived from this study should be largely applicable to ES.

The results of the study are provided in Table~\ref{Tab:find}. The first row indicated by \texttt{ref} is the reference model based on the conventional object detection task that the center-based representation is originally designed for. Thus, the representation is deemed effective for a particular task (a row in the table) if it achieves APs that are comparable or higher than those of \texttt{ref}. It can be clearly seen that the representation is effective for most of the tasks, except for the class-specific {\em stuff}-only and {\em thing}-{\em stuff} tasks which have much weaker performances. This finding suggests that the center-based representation is poorly effective only if both of these conditions are satisfied: (1) task is class-specific; (2) {\em stuffs} are involved. Furthermore, we find that the representation has a surprisingly strong compatibility with all the tasks with class-agnostic training, as demonstrated by their excellent APs. These two interesting findings provide us with a powerful support as to why using the center-based representation to handle both {\em things} and {\em stuffs} in a unified manner is the right choice for ES, but it is not necessarily the case for existing class-specific tasks that involve {\em stuffs}, such as panoptic segmentation.

In Table~\ref{Tab:find}, there is a significant drop in the $\text{AP}^{b}$ for \textit{stuff} when we move from class-agnostic training to class-specific training. Unlike object instances (\textit{thing}), 
a \textit{stuff} region often cover much larger areas of the image. Although center-based representation (which uses features from a single point of feature maps) provides a smaller receptive field and more local context, it works effectively for class-agnostic stuff mask prediction. We hypothesize that this is because all subregions (even for spatially distant subregions) within a stuff region tend to share similar colors and textures as the subregion around its center location. E.g., \textit{sky}, \textit{grass}, and \textit{water} usually share the same colors and textures within their respective regions. However, to accurately classify \textit{stuff} regions, the model has to take into account the semantics of other faraway neighboring regions (regardless \textit{thing} or \textit{stuff}), which requires a larger receptive field and more global context. Due to its localized nature, center-based representation performs poorly when a larger receptive field and more global context are needed, thus degrading the class-specific task performance.

\section{Method}
As mentioned in Sec.~\ref{sec:analysis}, all entities are represented by the center-based unified entity representation -- \textit{i.e.,} we represent each entity (including {\em stuffs}) by the {\em mass center} of its mask, similar to how objects are represented by their bounding-box centers \cite{tian2019fcos}. In this section, we first describe how we devise an {\em Entity segmentation framework} based on the unified center-based representation. Then, we introduce two novel modules: {\em global kernel bank} and {\em overlap suppression}, to exploit the requirements of ES in order to improve the segmentation quality of predicted entity masks.

\subsection{Segmentation Framework}
%Recall from the task format
As described in Sec.~\ref{sec:concept}, a non-overlapping entity ID prediction map $\mathbf{P}$ and a list of confidence score
map $\mathbf{S}$ are expected, given an input $\mathbf{I}$. Inspired by recent one-stage segmentation approaches~\cite{tian2020conditional,wang2020solov2,li2020fully,qi2020pointins}, we adopt a similar {\em generate-kernel-then-segment} pipeline. Specifically, with a single stage feature $\mathbf{X}_i \in \mathbb{R}^{H_i \times W_i \times C_i}$ from the $i$-th stage in Feature Pyramid Network (FPN)~\cite{lin2017feature}, four branches are added to handle the four output types required to perform ES: {\em entityness}, {\em centerness}, {\em localization}, and {\em kernel}.
%Here, entityness and centerness branches respectively provide a probability map $\mathbf{E}_i \in [0,1]^{H_i \times W_i}$ that indicates each pixel's probability of being an entity and a centerness map $\mathbf{C}_i \in [0,1]^{H_i \times W_i}$ that estimates centerness values \cite{tian2019fcos}.
Here, entityness and centerness branches respectively provide a probability map $\mathbf{E}_i \in [0,1]^{H_i \times W_i}$ and a centerness map $\mathbf{C}_i \in [0,1]^{H_i \times W_i}$. $\mathbf{E}_i$ indicates each pixel's probability of being an entity, while $\mathbf{C}_i$ estimates centerness values \cite{tian2019fcos}.
Localization branch is used to regress the bounding box offsets $\mathbf{L}_i$
of entities, which are used for efficient non-maximium suppression (NMS). For mask generation, we draw inspirations from dynamic kernel work \cite{jia2016dynamic,tian2020conditional,li2020fully,qi2020pointins} and employ a kernel branch to generate entity-specific dynamic kernel weights. 

\subsection{Global Kernel Bank}\label{sec:kernel_bank}
The use of dynamic kernels to generate entity masks assumes that the cues needed to segment different entities are dissimilar, but there exists many common properties ({\em e.g.} textures, edges) which are shared by different entities. This motivates us to leverage static kernel weights for training the mask head, alongside the dynamic kernel weights. The combination of dynamic and static kernel weights, referred to as \textit{global kernel bank}, allows the network to strike a good balance between learning entity-specific features and becoming aware of features that are commonly useful to many entities.

Global kernel bank consisting of three pairs of dynamic and static convolutions can be easily incorporated into our segmentation framework. To strengthen the interactions between dynamic and static kernels, we allow either dynamic or static kernels at each convolutional layer, resulting in seven possible network paths (minus the one with solely static kernels) in a 3-layer mask head, as presented in Fig.~\ref{fig:framework}. During training, we train with all seven paths simultaneously to minimize the mask prediction (Dice \cite{milletari2016fully}) loss with respect to ground truth $\mathbf{Y}$:
% -------------***************************--------------%
\begin{equation}
\mathcal{L}_{r}^{seg} = \lambda_r \times \mathrm{Dice}(\mathrm{Sigmoid}(\mathrm{MaskHead}(\mathbf{\tilde{F}}; \theta_r)), \mathbf{Y}),
\label{eqn:l_seg_r}
\end{equation}
% -------------***************************--------------%
where $r$ and $\lambda_r$ respectively denote the path's index and the path-specific hyperparameter that weights the loss. The default value of $\lambda=\{\lambda_r$\}, where $r=\{1,...,7\}$, are [0.25, 0.25, 0.25, 1, 1, 1, 1]. we note that the first three weights are 0.25, which are smaller than the other. That's because the convolution weights of the first layer would be better in dynamic format to satisfy the image feature wrapped with position embedding. We apply the same $\lambda$ for all experiments, unless specified otherwise. $\mathbf{\tilde{F}} \in R^{H/8 \times W/8 \times (C_{mask} + 2)}$ indicates the concatenation of backbone features (encoded for mask prediction) and relative coordinates~\cite{liu2018intriguing}. The kernels denoted by $\theta_r$ can be represented more granularly by $\theta_r^{v_{u}}$ to indicate its kernel type ($v \in \{s\text{:static},d\text{:dynamic}\}$) and convolutional layer index ($u \in \{0,1,2\}$) in MaskHead. Given $\theta_r^{v_{u}}$, the $\mathrm{MaskHead}$ for each path $r$ is defined as follows
	
\begin{equation}
\begin{aligned}
\mathrm{MaskHead}(\mathbf{\tilde{F}}; \theta_1) =& \otimes(\theta_{1}^{s_{0}},\otimes(\theta_{1}^{s_{1}},\otimes(\theta_{1}^{d_{2}}, \mathbf{\tilde{F}}))) \\
\mathrm{MaskHead}(\mathbf{\tilde{F}}; \theta_2) =& \otimes(\theta_{2}^{s_{0}},\otimes(\theta_{2}^{d_{1}},\otimes(\theta_{2}^{s_{2}}, \mathbf{\tilde{F}}))) \\
\vdots & \\
\mathrm{MaskHead}(\mathbf{\tilde{F}}; \theta_7) =& \otimes(\theta_{7}^{d_{0}},\otimes(\theta_{7}^{d_{1}},\otimes(\theta_{7}^{d_{2}}, \mathbf{\tilde{F}}))),
\end{aligned}
\end{equation}
\noindent where $\otimes(\cdot, \cdot)$ indicates the spatial convolution operation.

Despite the many paths trained, during inference, we find that using the path with solely dynamic kernels $\theta_7$ alone is sufficiently effective while being computationally efficient.

\subsection{Overlap Suppression}\label{sec:overlap_suppression}
The ES task (Sec.~\ref{fig:task_definition}) expects no overlapping masks. However, dynamic kernel-basel approaches tend to generate overlapping masks with high confidence due to the concept overlap among adjacent entities and independent losses being used for different kernels. Although postprocessing strategies \cite{kirillov2019panopticfpn, li2020fully} can resolve mask overlaps, they are driven by handcrafted heuristics which are less effective. Instead, we propose a module that encourages the model to learn to suppress the overlaps among the predicted entity masks. 

During training, there are $M$ number of sampled kernels and each of the $M$ kernels is assigned to one of the $N$ ground-truth entities according to the mask-based sample assignment strategy \cite{tian2020conditional}. This can be viewed as having $N$ clusters with one or more kernels within each cluster. Considering that the 7$^{\text{th}}$ path (all dynamics weights) is the only path used for inference, we only perform overlap suppression for this path. First, we obtain the \textit{representative} mask Q$^n$ of the $n$-th cluster by averaging all its masks generated via ${\theta}_7$:
    
\begin{equation}
\begin{aligned}
\mathbf{Q^n} =& \frac{\sum_{i \in \Omega(n)} 
                \mathrm{MaskHead}(\mathbf{\tilde{F}}; {\theta}_{7\vert i})}{
                \vert \Omega(n) \vert}
\end{aligned}
\end{equation}

% -------------***************************--------------%
\noindent where $\Omega(n)$ returns the set of kernel indices belonging to the $n$-th cluster and $\theta_{7 \vert i}$ indicates the $i$-th kernel in $\theta_7$. Given the representative entity masks $\mathbf{Q}$, we apply %a
pixelwise softmax function to induce a strong suppression of non-maximal entities in the pixel-wise mask prediction. To achieve overlap suppression, we adopt a separate training loss similar to $\mathcal{L}_{r}^{seg}$ (Eq.~\ref{eqn:l_seg_r}):
% -------------***************************--------------%
\begin{equation}
\mathcal{L}_{o}^{seg} = \mathrm{Dice}(\mathrm{Softmax}(\mathbf{Q}), \mathbf{Y}),
\end{equation}
	
Note that the pixels without annotations are ignored in ${L}_{o}^{seg}$.

\subsection{Training and Inference}
In the training stage, the segmentation network is trained with the overall loss defined as: 
% -------------***************************--------------%
\begin{equation}
\mathcal{L} = \mathcal{L}^{det} + \mathcal{L}_{o}^{seg} + \mathcal{L}_{R}^{seg} =  \mathcal{L}^{det} + \mathcal{L}_{o}^{seg} + \sum_{r} {\mathcal{L}_{r}^{seg}}.
\end{equation}
% -------------***************************--------------%

In the inference stage, we first sort all entity detections according to their (aggregated) confidence scores: $\sqrt{\mathbf{C}_i \text{(centerness)} \times \mathbf{E}_i \text{(entityness probability)}}$, and their corresponding boxes are obtained from the localization branch and then processed with box-level NMS. After the duplicate removal, each remaining entity is encoded into an entity-specific dynamic kernel, resulting in $N$ kernels for $N$ entities. With the generated dynamic kernels and encoded features shown in Fig.~\ref{fig:framework}, the segmentation masks $\mathbf{\hat{Y}} \in \mathbb{R}^{N \times H_3 \times W_3}$ of $N$ entities are produced by a sequence of convolutions directly. Finally, the final non-overlapping prediction map $\mathbf{P}$ is acquired by choosing the entity ID with the maximum confidence score at each pixel ~\cite{kirillov2019panopticfpn}.

\section{Experiments}
\subsection{User study on the definition of entity}
\label{subsec:entity}
\begin{figure}[t!]
\centering
\includegraphics[width=\linewidth]{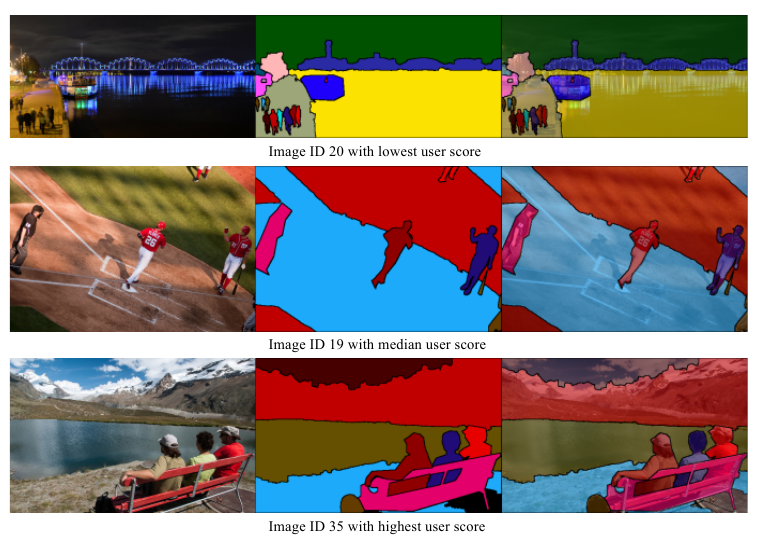}
\caption{The visualization of images and their corresponding entity annotations, which are used for our user study.}
\label{fig:user_study_2}
\end{figure}

\begin{figure*}[t!]
\centering
\includegraphics[width=\linewidth]{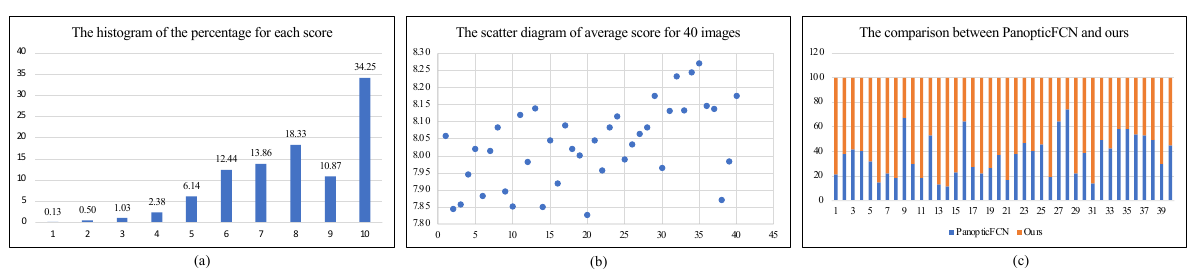}
\caption{The statistical visualizations of the data from our user and survey studies. \textbf{(a)} The histogram 
here represents the distribution of the users based on their given scores in the entity definition user study. \textbf{(b)} The scatter diagram that showcases the averaged scores (averaged across all users) for each of the 40 images. \textbf{(c)} Each horizontal bar here indicates the proportions of votes given by the survey participants to PanopticFCN~\cite{li2020fully} (blue) and {\em ours} (orange) on each of the 40 images. }
\label{fig:user_study}
\end{figure*}

We conducted a user study in which there were 480 individuals who were identified as Adobe Photoshop users who regularly used the software for image manipulation/editing. In this user study, we randomly selected 40 images of COCO dataset and provided the users a visualization of the {\em entity} ground truths drawn with distinct colors. For each image, we asked each user about his/her degree of satisfaction about treating semantically-meaningful and -coherent segments as entities, with respect to their relevance to and suitability for image manipulation/editing applications. 
The satisfactory scores aggregated from all users for the individual images. We find that the average score of each image is large than 7.8 on the condition that the maximum score is 10. Most of the selected images' scores are larger than 6.0, This confirms that the users are highly satisfied with our task's definition of {\em entity} (and the lacks of category labels and {\em thing-stuff} separation) in the context of image manipulation/editing. To better present the user study's findings, we summarize the data in Figure~\ref{fig:user_study}(a) \&(b). The image IDs with the minimum, median and maximum user scores are \textsc{20}, \textsc{19}, and \textsc{35}. We show these images and their corresponding entity annotations in Fig.~\ref{fig:user_study_2}.

\subsection{Main Evaluation}
We perform the main evaluation studies through the single-dataset setting on COCO dataset~\cite{lin2014microsoft}. Following the common practice, we train our models with 118,287 train images and reported results on the 5,000 validation images. Unless specified, we train our network with ImageNet-pretrained ResNet-50 \cite{he2016res} backbone using batch size $16$ for 12 epochs. The longer edge sizes of the images is $1333$.  The shorter edge sizes of the images is random sampled from $640$ to $800$ with stride $32$. We decay the learning rate with $0.1$ after $8$ and $11$ epochs respectively.

We note that our baseline model is the CondInst~\cite{tian2020conditional}, a popular instance segmentation model. In our baseline, we preserve all components of CondInst except that we set the number of categories to 1. Furthermore, we do not modify any parts of the segmentation head architecture of CondInst.

\begin{table*}[t!]
\caption{Ablation studies. $\circ$ and $\checkmark$ respectively indicate whether a particular component is ablated or in used. \textbf{(a)}: \textbf{Proposed two modules}. The impacts of global kernel bank and overlap suppression modules. \textbf{(b)}: \textbf{Overlap suppression}. An ablation study of the scoring activation function in the module. \textbf{(c)}: \textbf{The module of global kernel bank.} Each of ``1,2,...,7''corresponds to the $r$-th path in global kernel bank, as in Fig.~\ref{fig:framework}. \textbf{(d)}: \textbf{High-Performance Regime}. The performance of our models enhanced by stronger backbones and a longer training duration. The ``DCNv2'' refers to deformable convolution v2~\cite{zhu2019deformable}.}
\label{Tab:ablation1}
\begin{minipage}{\textwidth}
        \begin{minipage}[t]{0.2\textwidth}
            \centering
            \small
            \setlength{\tabcolsep}{2pt}
            \begin{tabular}{cc|c}
            \toprule
            $\mathcal{L}_{R}^{seg}$ & $\mathcal{L}_{o}^{seg}$ & AP$_\text{e}^\text{m}$ \\ \midrule
            $\circ$ & $\circ$ & 28.3  \\ \midrule
            $\checkmark$ & $\circ$ & 29.3  \\ \midrule
            $\circ$ & $\checkmark$ & 29.1  \\ \midrule
            $\checkmark$ & $\checkmark$ & \textbf{29.8} \\ \bottomrule
            \end{tabular}
        
        (a)
        \end{minipage}
        \begin{minipage}[t]{0.2\textwidth}
        \centering
        \small
        \setlength{\tabcolsep}{2pt}
        \begin{tabular}{c|c|c}
            \toprule
            Softmax & Sigmoid & AP$_\text{e}^\text{m}$ \\ \midrule
            $\circ$ & $\circ$ & 28.3 \\ \midrule
            $\circ$ & \checkmark & 28.6  \\ \midrule
            \checkmark &$\circ$ &  \textbf{29.1}\\ \midrule
            \checkmark &\checkmark &  29.0\\ \midrule
        \end{tabular}
        
    (b)
    \end{minipage}
    \begin{minipage}[t]{0.3\textwidth}
            \centering
            \small
            \setlength{\tabcolsep}{2pt}
            \begin{tabular}{c|c|c|c|c|c|c|c}
            \toprule
            1 & 2 & 3 & 4 & 5 & 6 & 7 & $\text{AP}_\text{e}^\text{m}$ \\ \midrule
            $\circ$ & $\circ$ & $\circ$ & $\circ$ & $\circ$ & $\circ$ & \checkmark & 28.3 \\ \midrule 
            $\circ$ & $\circ$ & $\circ$ & $\circ$ & $\circ$ & \checkmark & \checkmark & 28.9 \\ \midrule  
            $\circ$ & $\circ$ & $\circ$ & $\circ$ & \checkmark & \checkmark & \checkmark & 29.1 \\ \midrule 
           \checkmark & \checkmark & \checkmark & \checkmark & \checkmark & \checkmark & \checkmark &\textbf{29.3} \\ \bottomrule 
           \end{tabular}
        
        (c)
        \end{minipage}
        \begin{minipage}[t]{0.2\textwidth}
        \centering
        \small
        \setlength{\tabcolsep}{2pt}
        \begin{tabular}{c|c|c}
            \toprule
            Backbone & Epochs & $\text{AP}_\text{e}^\text{m}$ \\ \midrule
            R-50 & 12 & 29.8 \\ \midrule
            R-50 & 36 & 31.8 \\ \midrule
            R-101 & 36 & 33.2 \\ \midrule
            R-101-DCNv2 & 36 & \textbf{35.5} \\ \midrule
        \end{tabular}
        (d)
        \end{minipage}
        
\end{minipage}
\end{table*}

\textbf{Proposed two modules.} Table~\ref{Tab:ablation1}(a) summarizes the performance improvements introduced by the two proposed modules to the baseline. The baseline is a simple segmentation framework that uses only the loss $\mathcal{L}_{0}^{seg}$ in mask branch, and it obtains 28.3 AP$_{\text{e}}^\text{m}$. The incorporation of $\mathcal{L}_{R}^{seg}$ and $\mathcal{L}_{o}^{seg}$ improves it by 1.0, 0.8 AP$_{\text{e}}^\text{m}$, respectively, demonstrating the effectiveness of the proposed modules. The last line shows that using these two modules achieves an even greater improvement of 1.5 AP$_{\text{e}}^\text{m}$. This indicates that the two proposed modules are complementary with each other. Note that these two modules can bring consistent improvements even with a large state-of-the-art backbone. Particularly, with a large (L) Swin Transformer backbone, the two modules still provide a good performance gain of 1.4 AP$_{\text{e}}^\text{m}$.

\textbf{Overlap suppression.} 
Table~\ref{Tab:ablation1}(b) provides the ablation study on the activation function used in $\mathcal{L}_{o}^{seg}$. Compared to the baseline without any activation function, Sigmoid and Softmax respectively bring 0.3 and 0.8 AP$_{\text{e}}^\text{m}$ improvements. Compared to Sigmoid, Softmax forces the network to suppress the mask prediction of non-maximal entities at every pixel and this helps to suppress mask overlaps more effectively. In the last row of Table~\ref{Tab:ablation1}(c), we use two ${L}_{o}^{seg}$ losses based on Sigmoid and Softmax separately, with a loss weight of 0.5 for each. 

\textbf{Global kernel bank.}
Table~\ref{Tab:ablation1}(c) presents the ablation study on the choice of the mask head paths used in the training stage. The first row is dynamic-kernel baseline that obtains 28.3 AP$_{e}^{m}$, using only the dynamic kernel weights for all layers. Simply using static weights in the last one or two layers improves the AP$_{\text{e}}^\text{m}$ by 0.6 and 0.8, as shown in the second and third rows. The best performance is achieved from including all the seven paths  ($\theta_1,\theta_2,...\theta_7$). This allows the model to achieve the right balance between becoming aware of the common properties of entities (\textit{e.g.,} textures or edges) and preserving a decent entity discriminability.

In the table\ref{tab::global_inference}, we show the results of an ablation study on the choice of paths for the global kernel bank module during inference. We find that using all paths does not provide performance advantage over using just the last path. Given the effectiveness of using just the last path and its greater computational efficiency, we decide to perform inference using just the last path with three dynamic convolution weight layers.

\begin{table}[h!]
    \centering
    \caption{The ablation study on the choice of path for global kernel bank module during inference}.
    \label{tab::global_inference}
    \begin{tabular}{c|c}
       Paths & AP$_\text{e}^\text{m}$ \\ \hline
       Last Path & 29.8 \\ \hline
       All Paths & \textbf{30.0} \\ \hline
    \end{tabular}
    
\end{table}

\textbf{The consistency among Table~\ref{Tab:ablation1}.} The results in tables~\ref{Tab:ablation1}(a), \ref{Tab:ablation1}(b), and \ref{Tab:ablation1}(c) are actually consistent with each other. The first row in each of these three tables corresponds to our baseline model, which is essentially CondInst~\cite{tian2020conditional} without any structural or architectural modifications. In Table 3(a), the second row shows the performance gain from adding our proposed \textit{global kernel bank to the baseline model}, while the third row shows the gain from adding the \textit{overlap suppression} module. The performance of these two rows corresponds to the best versions in table~\ref{Tab:ablation1}(c) and \ref{Tab:ablation1}(b) respectively, where the other rows of them ablates different choices in each module. When we perform ablation study on either one of the modules in Table~\ref{Tab:ablation1}(c) or \ref{Tab:ablation1}(b), we leave out the other proposed module (e.g., the models in Table 2(b) do not use make use of \textit{global kernel bank}). By removing the influence of the other module, we can have a better understanding of the module being evaluated. Finally, ours (Condinst~\cite{tian2020conditional} with the two proposed modules) obtains the best performance as shown by the last row of Table~\ref{Tab:ablation1}(a).

Moreover, we independently ablate the proposed two modules through a more standard experimental setting that includes the other module in the model. Table~\ref{tab:ablation_overlap_global} and \ref{tab:ablation_global_overlap} demonstrate similar improvement trends of the proposed two modules as compared to Table~\ref{Tab:ablation1}(b) and (c) which are based on an alternative setting that leaves out the other module in the model.

\textbf{High-performance regime.} 
Table~\ref{Tab:ablation1}(d) shows the performance of our proposed method in the high-performance training regime: stronger backbones and a longer training duration. With the longer training duration, we carry out training with a batch size of 16 for 36 epochs. We decay the learning rate with 0.1 factor at the 33-th and 35-th epochs. The models trained with our method benefit from the various techniques of high-performance training regime. The strongest one with ResNet-101 and Deformable Convolution v2~\cite{zhu2019deformable} obtains 35.5 AP$_{\text{e}}^\text{m}$, and thus it is used for the cross-dataset visualization in the next section.

In the ablation study experiments represented by Table~\ref{Tab:ablation1}(a), \ref{Tab:ablation1}(b), and \ref{Tab:ablation1}(c), the number of our training epoch is 12, which is consistent with the standard ablation study setting of object detection and instance segmentation papers~\cite{lin2017feature,he2017mask,liu2018path}. To further prove the effectiveness of our proposed method under a long training schedule (36 epochs), we observe a similar performance boost from our proposed modules in Table~\ref{tab:ablation_36epochs}.

\begin{table}[t!]
    \centering
    \caption{The ablation study on the proposed modules in 36 training epochs, where the models could obtain the full convergence.}
    \label{tab:ablation_36epochs}
    \begin{tabular}{c|c}
       Model & AP$_\text{e}^\text{m}$ \\ \hline
       class-agnostic Condinst~\cite{tian2020conditional} (baseline) & 30.4 \\ \hline
       ours with two proposed modules & 31.8 \\ \hline
    \end{tabular}
\end{table}

\begin{table}[t!]
    \centering
    \caption{The ablation study on the overlap suppression module of the model wrapped with the global kernel bank module.}
    \label{tab:ablation_overlap_global}
    \begin{tabular}{c|c|c}
    Softmax & Sigmoid & AP$_\text{e}^\text{m}$ \\ \hline
    $\circ$ & $\circ$ & 29.3 \\ \hline
    $\circ$ & \checkmark & 29.5  \\ \hline
    \checkmark &$\circ$ & \textbf{29.8} \\ \hline
    \checkmark &\checkmark &  \textbf{29.8}\\ \hline
    \end{tabular}
\end{table}

\begin{table}[t!]
    \centering
    \caption{The ablation study on the global kernel bank module of the model wrapped with the overlap suppression module.}
    \label{tab:ablation_global_overlap}
    \begin{tabular}{c|c|c|c|c|c|c|c}
    1 & 2 & 3 & 4 & 5 & 6 & 7 & $\text{AP}_\text{e}^\text{m}$ \\ \hline
    $\circ$ & $\circ$ & $\circ$ & $\circ$ & $\circ$ & $\circ$ & \checkmark & 29.1 \\ \hline 
    $\circ$ & $\circ$ & $\circ$ & $\circ$ & $\circ$ & \checkmark & \checkmark & 29.6  \\ \hline  
    $\circ$ & $\circ$ & $\circ$ & $\circ$ & \checkmark & \checkmark & \checkmark & 29.7 \\ \hline 
   \checkmark & \checkmark & \checkmark & \checkmark & \checkmark & \checkmark & \checkmark &\textbf{29.8} \\ \hline 
   \end{tabular}
\end{table}

\begin{table*}[t!]
\caption{Cross-dataset evaluation via %\textbf{(a)} and \textbf{(b)}:
the comparisons with \textbf{(a) PanopticFCN~\cite{li2020fully}} %in
and \textbf{(b) PanopticDETR~\cite{carion2020end}}. 
Each column and row represent the training and validation dataset we use, respectively. "CO+A" means we use both COCO and ADE20K training datasets. 
Each cell element in Table (a)/(b) corresponds to the results of: (1) PanopticFCN/DETR with class-specific training; (2) with class-agnostic training; (3) Ours with 12 epochs in (a) and 36 epochs in (b). \textbf{(c)}: The results of using stronger backbones to train on the merged COCO+ADE20K dataset.}
\label{Tab:cross_inference}
\begin{minipage}{\textwidth}
    \begin{minipage}[t]{0.44\textwidth}
        \centering
        \small
        \setlength{\tabcolsep}{2pt}
        \begin{tabular}{c|c|c|c}
        \toprule
        Dataset & COCO & ADE20K & CO+A \\ \midrule
        COCO & 24.5/26.5/\textbf{29.8} & 15.4/16.6/\textbf{19.4} & 25.8/27.8/\textbf{30.6} \\ \midrule
        ADE20K & 15.2/17.5/\textbf{20.1} & 20.6/21.3/\textbf{24.1} & 23.8/24.9/\textbf{26.3} \\ \midrule
        CO+A & 22.1/23.8/\textbf{27.0} & 16.6/17.8/\textbf{20.5} & 24.9/26.6/\textbf{28.9} \\
        \bottomrule
        \end{tabular}
    
    (a)
    \end{minipage}
    \begin{minipage}[t]{0.23\textwidth}
    \centering
    \small
    \setlength{\tabcolsep}{2pt}
    \begin{tabular}{c|c}
        \toprule
        Dataset & COCO \\ \midrule
        COCO & 24.8/29.2/\textbf{31.8}   \\ \midrule
        ADE20K & 16.4/18.9/\textbf{21.1} \\ \midrule
        CO+A   & 22.9/26.3/\textbf{28.7} \\ \bottomrule
    \end{tabular}
    
(b)
\end{minipage}
\begin{minipage}[t]{0.3\textwidth}
    \centering
    \small
    \setlength{\tabcolsep}{2pt}
    \begin{tabular}{c|c|c}
        \toprule
        Transformer & COCO & ADE20K \\ \midrule
        Swin-T~\cite{liu2021swin} & 35.0 & 33.2 \\ \midrule
        Swin-L~\cite{liu2021swin} & \textbf{38.9} & \textbf{37.0} \\ \midrule
        MiT-B5~\cite{xie2021segformer} & 37.4 & 35.6 \\ \bottomrule
    \end{tabular}
    
(c)
\end{minipage}
    
\end{minipage}
\end{table*}

\begin{table}[t!]
\centering
\small
\caption{\textbf{Comparison with panoptic segmentation methods}. The different train-test settings are: (S-S) class-specific training \& class-specific testing; (S-A) class-specific training \& class-agnostic testing; (A-A) class-agnostic training \& class-agnostic testing. The PQ and AP$_\text{e}^\text{m}$ are the evaluation metrics of panoptic segmentation and ES. Note that all models here are trained until full convergence by using long training schedules\protect\footnotemark[1].}
\label{Tab:ablation2}
\begin{tabular}{c|c|c|c}
\toprule
Model & \makecell[c]{(S-S)\\PQ} & \makecell[c]{(S-A)\\AP$_\text{e}^\text{m}$} & \makecell[c]{(A-A)\\AP$_\text{e}^\text{m}$}   \\ \midrule
PanopticFPN~\cite{kirillov2019panopticfpn} & 41.5 & 25.2 & - \\ \midrule
PanopticFCN~\cite{li2020fully}  & 43.6 & 26.5 & 28.5 \\ \midrule  
PanopticDETR~\cite{carion2020end} & 43.4 & 26.8 & 29.2 \\ \midrule
KNet~\cite{zhang2021k} & 44.8 & 28.2 & 30.3 \\ \midrule
MaskFormer~\cite{cheng2021per} & 46.5 & 29.4 & 31.2 \\ \midrule
Ours & - & - & \textbf{31.8} \\ \bottomrule
\end{tabular}
\end{table}

\textbf{Comparison with panoptic segmentation methods.} In Table~\ref{Tab:ablation2}, we compare our segmentation method with PanopticFPN \cite{kirillov2019panopticfpn}, and DETR~\cite{carion2020end}, PanopticFCN~\cite{li2020fully}, and MaskFormer~\cite{cheng2021per}. As shown in the second and third column, the numbers of AP$_{\text{e}}^\text{m}$ are also smaller than PQ's under the similar class-specific training models. This is because PQ merely uses a single IoU threshold of $0.5$ to determine true positives, while our stricter AP$_{\text{e}}^\text{m}$ considers a range of IoU thresholds ranging from $0.5$ to $0.95$ with a step size of $0.05$. In the last two columns, the models with class-agnostic training consistently perform better in AP$_{\text{e}}^\text{m}$ than the class-specific ones. This observation reaffirms the importance of our entity segmentation task that prioritizes mask quality over semantic classification. Our model obtains the best performance by virtue of our two elaborately-designed modules that effectively exploit the unique class-agnostic and non-overlapping requirements of the task. Moreover, owing to the unified center-based representation and convolutional architecture, our method merely requires 36 training epochs to train the model to full convergence, compared to the hefty 300 epochs needed by Transformer-based methods~\cite{carion2020end,cheng2021per}. Thus, our method is significantly more accessible to many of the researchers and practitioners who have limited compute resources. Besides, Table~\ref{Tab:ablation2} shows that the two mainstream \textit{decoder} approaches to learn a unified representation for entities: dynamic convolutional weights ~\cite{li2020fully,zhang2021k} and Transformer with learnable queries~\cite{carion2020end,cheng2021per} can obtain comparable performance for Entity Segmentation. Here, we refer to \textit{encoder} as the model's backbone and \textit{decoder} as the model's prediction heads.

We note that all evaluated models within the same table use the same encoder (backbone and neck) unless specified otherwise. Compared to models with cascaded processing such KNet~\cite{zhang2021k} and MaskFormer~\cite{cheng2021per}, our proposed method introduces fewer computational overhead in the decoder part. In particular, for the same 800px image, our decoder requires 104.83 GFLOPs, while MaskFormer~\cite{cheng2021per} and KNet~\cite{zhang2021k} require 126.53 and 176.25 GFLOPS respectively. To further validate the effectiveness of our proposed modules, we incorporate the two modules to KNet and achieves 1.0\% AP$^\text{e}_\text{m}$ performance improvement. Besides, we enhance MaskFormer with our proposed \textit{overlap suppression} module and observe a 0.6{\%} gain in AP$_\text{e}^\text{m}$. Since MaskFormer is a query-based Transformer approach that does not rely on dynamic kernels, we neither combine nor evaluate MaskFormer with \textit{global kernel bank}.

Also, we compare our baseline model with the PanopticFCN~\cite{li2020fully} model trained using class-specific and -agnostic training strategy. All models are trained with 36 epochs for full convergence. Besides the AP$^{\text{e}}$ metric that we use to demonstrate the effectiveness of class-agnostic training for PanopticFCN \& DETR in Table~\ref{Tab:cross_inference} (a)(b), the Table~\ref{tab::em} shows that our proposed method still improves segmentation quality (SQ) over the class-specific PanopticFCN.

\begin{table}[h!]
    \centering
    \small
    \caption{The comparison between PanopticFCN~\cite{li2020fully} and ours under the SQ metric.}
    \label{tab::em}
    \begin{tabular}{c|c|c}
       Method & Train & SQ \\ \hline
       \multirow{2}*{PanopticFCN~\cite{li2020fully}} & class-specific & 78.6 \\ \cline{2-3}
       & class-agnostic & 79.3 \\ \hline
       ours & class-agnostic & \textbf{79.9} \\ \hline
    \end{tabular}
\end{table}

\textbf{The representations in stuff.} In the table~\ref{table:rep_stuff}, we show the result of representing entities using points located within entity mask regions (i.e., stuff regions). The points are selected in such a way that they are closest to their corresponding mass centers. However, such a representation brings no improvement. By following the approach of convolution-based panoptic segmentation model PanopticFCN \cite{li2020fully} that samples multiple points of stuff region for feature aggregation, we obtain a better performance at the cost of 2.1 times computation overhead. Therefore, for a good tradeoff between efficiency and accuracy, we still opt for using single mass centers to represent entities. These other two representations do not offer any performance advantages for \textit{things}.

\begin{table}[t!]
    \centering
    \small
    \caption{The ablation study on the representations of the stuff region.}
    \label{table:rep_stuff}
    \begin{tabular}{c|c}
    \toprule
       Center Type for Stuff & AP$_\text{e}^\text{m}$ \\ \midrule
       Single point (baseline) & 39.4 \\ \midrule
       Single Point (constrained within stuff region) & 39.4 \\ \midrule
       Nine Points (from stuff region) & \textbf{40.2} \\ \bottomrule
    \end{tabular}
\end{table}

\subsection{Cross-Dataset Evaluation}
To demonstrate the generalization advantage of our entity segmentation task, we consider the setting where the evaluation set comes from a dataset different than the one/ones used for training. Here, we experiment with COCO and ADE20K. All these two datasets are converted into the ES format. For fair comparisons, all models are trained with pre-sampled 138,497 images (118,287 from COCO and 20,210 from ADE20K)
regardless which datasets are used. PanopticFCN and DETR are trained for 12 and 300 epochs respectively, following their original implementations. Our models with Transformer backbones~\cite{liu2021swin,xie2021segformer} are trained for 36 epochs.

\footnotetext[1]{The convolution-based (PanopticFPN, PanopticFCN and ours) and the transformer-based frameworks (DETR and MaskFormer) adopt 36 and 300 training epochs, respectively. They all share the same ResNet50 backbone.}

\begin{figure*}[t!]
\begin{center}
\includegraphics[width=\linewidth]{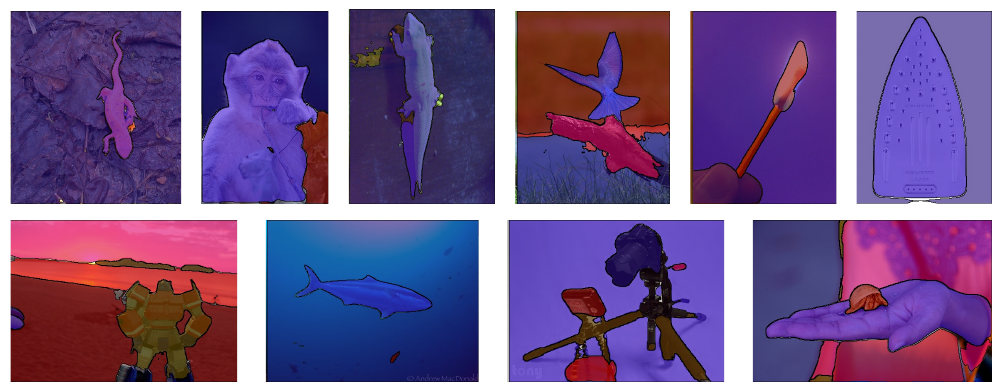}
\end{center}
\caption{The illustration of our model's generalization ability. We train the model in COCO dataset but apply it to ImageNet images. Note that most segmented entities are outside of COCO classes.}
\label{fig:generalization}
\end{figure*}

In our proposed entity segmentation (ES) task, the concept of entity unifies both things and stuffs, without differentiating between them. According to the ES task definition, we do not have access to information whether a particular entity belongs to thing or stuff. Therefore, we cannot adopt different strategies for thing and stuff and regard all entities as thing when applying existing panoptic segmentation models to ES. Furthermore, existing panoptic segmentation methods that adopt different strategies to handle stuff predictions using a fixed number of output channels (\textit{i.e.}, assuming a fixed number of stuff categories), in a similar fashion as semantic segmentation methods \cite{xiong2019upsnet, li2020fully}. However, in ES, we neither have a fixed set of stuff categories nor we know beforehand how many stuff masks there are for incoming input images. Moreover,
having a unified entity concept helps to avoid the thing-stuff annotation ambiguity, \textit{e.g.}, ``window" belongs to stuff in 
COCO~\cite{lin2014microsoft} dataset, while it belongs to thing in 
ADE20K~\cite{zhou2017scene} dataset.

\textbf{Quantitative evaluation.}
Table~\ref{Tab:cross_inference}(a) and (b) shows the performance comparisons with PanopticFCN~\cite{li2020fully} and DETR~\cite{carion2020end} in the cross-dataset setting. From those two tables, it can be clearly seen that our models generalizes better in the cross-dataset setting. %Except that, we find some confusing phenomenon need to explain. 
In Table~\ref{Tab:cross_inference}(a), the model trained on only COCO dataset obtains 29.8 and 20.1 AP$_{\text{e}}^\text{m}$ on COCO and ADE20K validation sets, signifying the superior generalization advantages of our task and proposed segmentation framework.

However, the COCO-trained model performs worse on ADE20K ($29.8$$\rightarrow$$20.1$ AP$_{\text{e}}^\text{m}$). There are two reasons: (1) certain classes common to COCO and ADE20K are distinguished as {\em thing} and {\em stuff} differently; (2) ADE20K has more noisy annotations. Thus it may unreasonably penalize the good mask predictions from our model, which conflict with the annotations of other datasets due to large annotation gaps. Despite that, the qualitative results of ADE20K in Fig.~\ref{vis_various_dataset} indicate that our model achieves a reasonably good performance. Also, there is a big performance gap on COCO validation step between using COCO and ADE20K for training. This is caused by ADE20K's low number (20,210) of training samples that are insufficient to train the model well. 

One interesting aspect of our proposed task and framework is that multiple datasets of different domains can be directly combined to form a large training dataset. This advantage still remains when the models are trained to full convergence, as shown in Table\ref{Tab:cross_inference}(b). The aforementioned COCO$\rightarrow$ADE20K problem 
is easily solved by combining COCO and ADE20K (COCO+A) for training without resolving label conflicts. In that case, the model achieves the best overall performance on all datasets. Furthermore, Table~\ref{Tab:cross_inference}(c) shows that using stronger encoder Transformer backbones can further bridge the performance gaps between the two.

In Table~\ref{Tab:cross_inference}(b), we keep the network architectures between different models (rows) as identical as possible. A network architecture for panoptic or entity segmentation generally consists of an encoder and a decoder, where the encoder is responsible for extracting visual features and the decoder decodes the visual features to generate predictions. We would like to point out that the decoder network structure is tightly coupled with the adopted representation and thus cannot be easily separated from the choice of representation. \textit{E.g.}, CondInst \cite{tian2020conditional} that uses center representation requires a convolution-based decoder to generate dynamic weights. A query-based Transformer decoder cannot be directly integrated to CondInst for it to adopt query representation. The same applies to DETR \cite{carion2020end} in which its query-based Transformer decoder cannot be directly adapted to support center-based representation. When we experimentally compare our center-based and query-based representation methods, we make sure that the same encoder (ResNet-50 backbone) is used by all methods, but their decoders may vary depending on the type of representation adopted.

\subsection{Visualization results and survey study for comparison between PanopticFCN and {\em ours}}
\begin{figure}[t!]
\begin{center}
\includegraphics[width=0.95\linewidth]{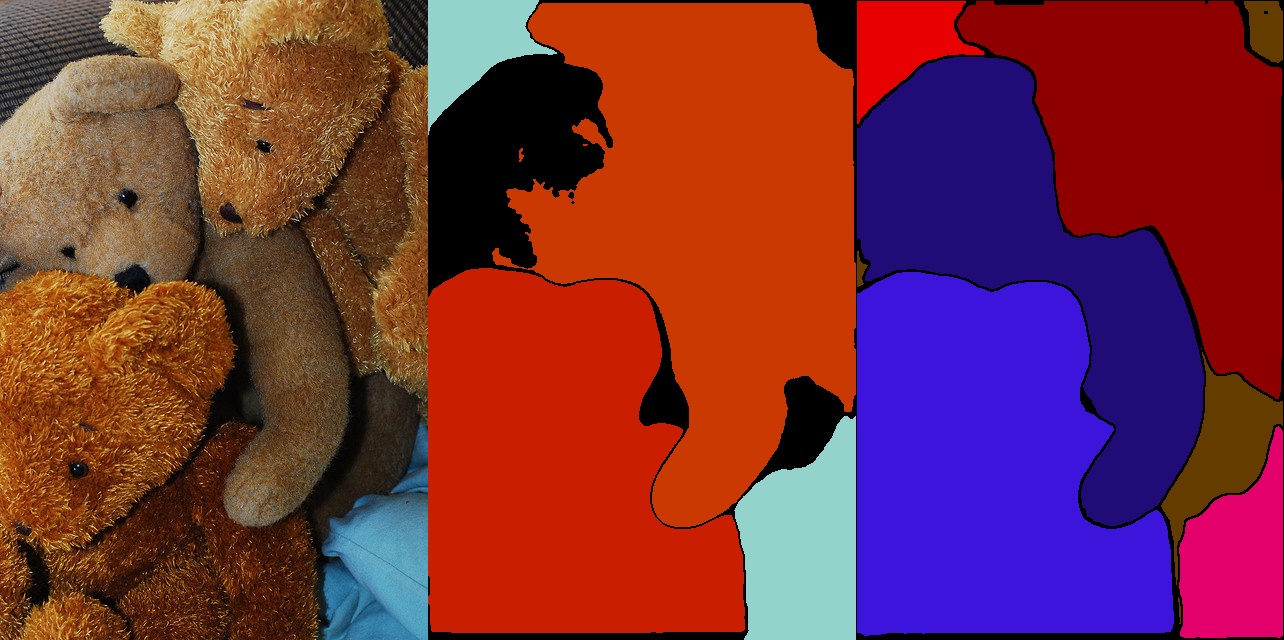} 
\vspace{0.01in}
\includegraphics[width=0.95\linewidth]{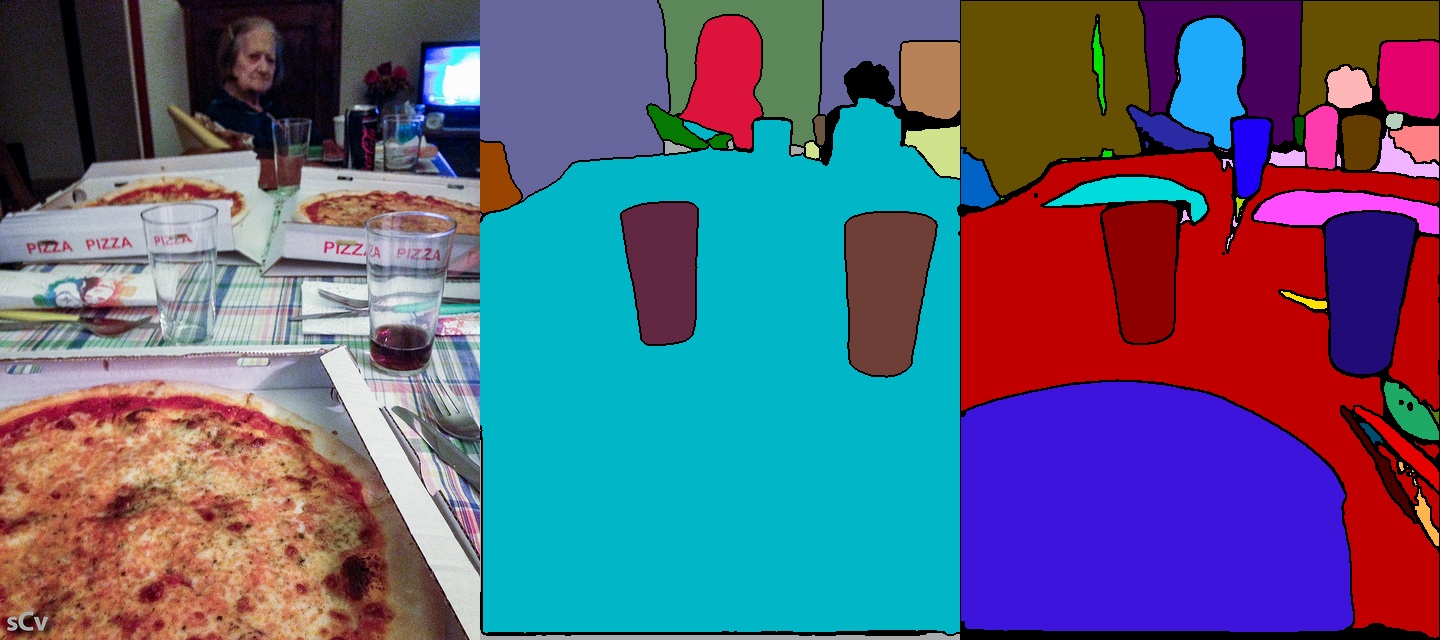}
\vspace{0.01in}
\includegraphics[width=0.95\linewidth]{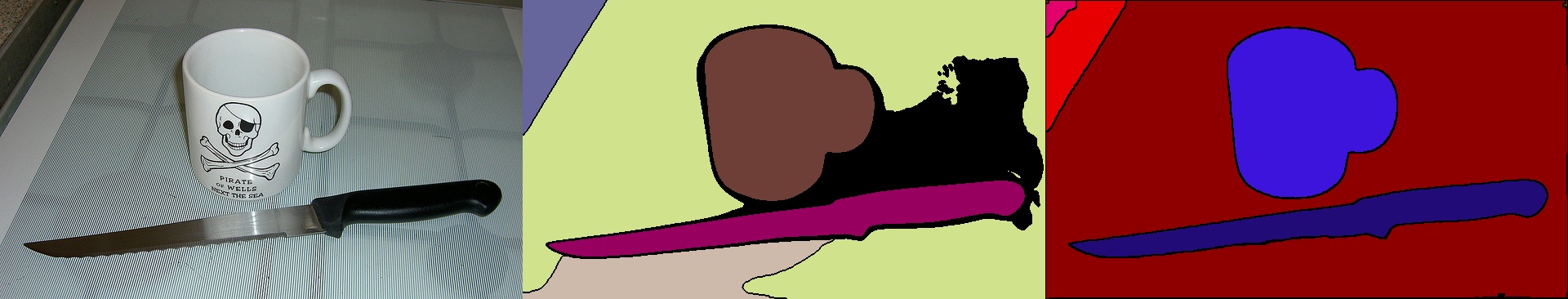}
\vspace{0.01in}
\includegraphics[width=0.95\linewidth]{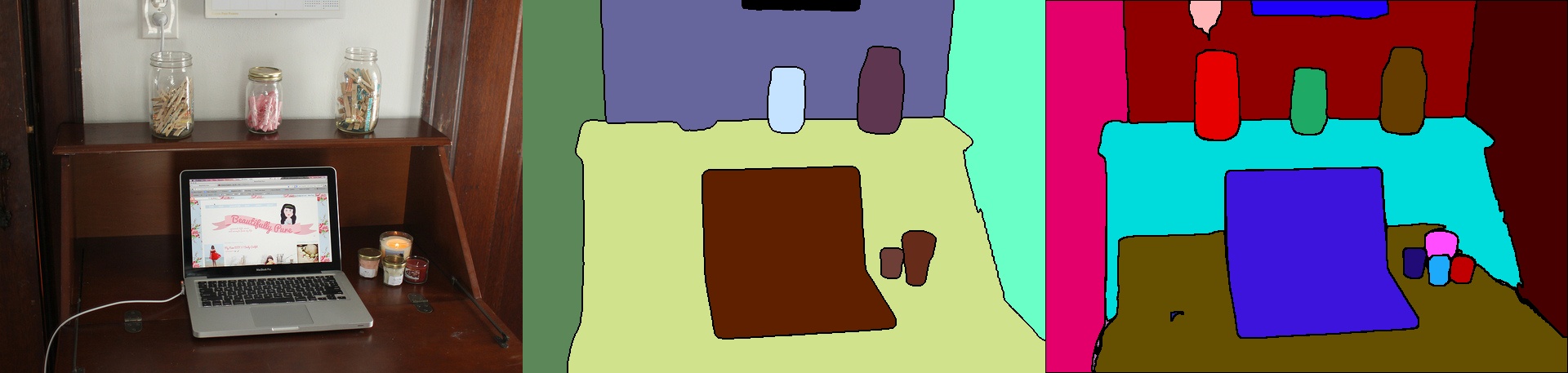}
\end{center}
\begin{tabular}{ccc}
\hspace{0.05\columnwidth}(1) The input & \hspace{0.02\columnwidth}(2) Panoptic FCN. & \hspace{0.005\columnwidth}(3) Our results. \\
\end{tabular}
\caption{Visual comparisons on COCO dataset with PanopticFCN and ours.}
\label{fig_cmp1}
\end{figure}

The segmentation results produced by a recent panoptic segmentation method and our entity segmentation framework are 
visually compared in Figure~\ref{fig_cmp1}. For panoptic segmentation, PanopticFCN with ResNet101~\cite{he2016res} backbone and Deformable Convolution v2~\cite{zhu2019deformable} is used as a strong contender for comparison here, as it achieves state-of-the-art performance among all publicly-released codes. Furthermore, we randomly selected 40 images (also included in Figure~\ref{fig_cmp1} and conducted a survey study of 480 people about their preference for the segmentation results produced by PanopticFCN and {\em ours}. As shown in Figure~\ref{fig:user_study}(c), the survey participants, to a great extent, preferred the segmentation results of our entity segmentation framework over PanopticFCN's.

\subsection{Qualitative results to open world.} Fig.~\ref{fig:generalization} shows the our model's qualitative results on ImageNet. The model used here is based on the R-101-DCNV2 backbone described in Table~\ref{Tab:ablation1}(d) and trained only on COCO dataset. Some rare entities are accurately segmented even though they never appear in the training set.

\begin{figure}[t!]
\begin{center}
\includegraphics[width=0.95\linewidth]{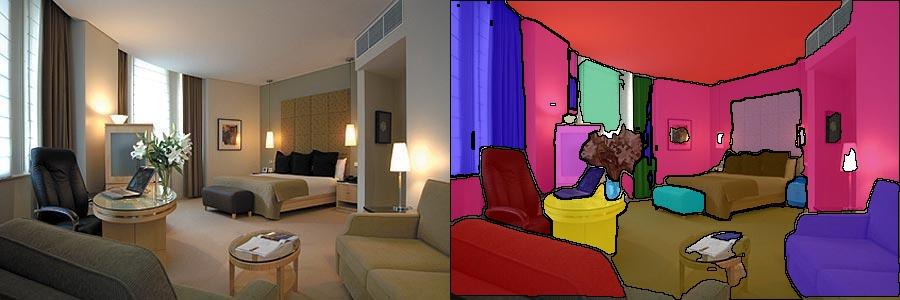} 
\vspace{0.01in}
\includegraphics[width=0.95\linewidth]{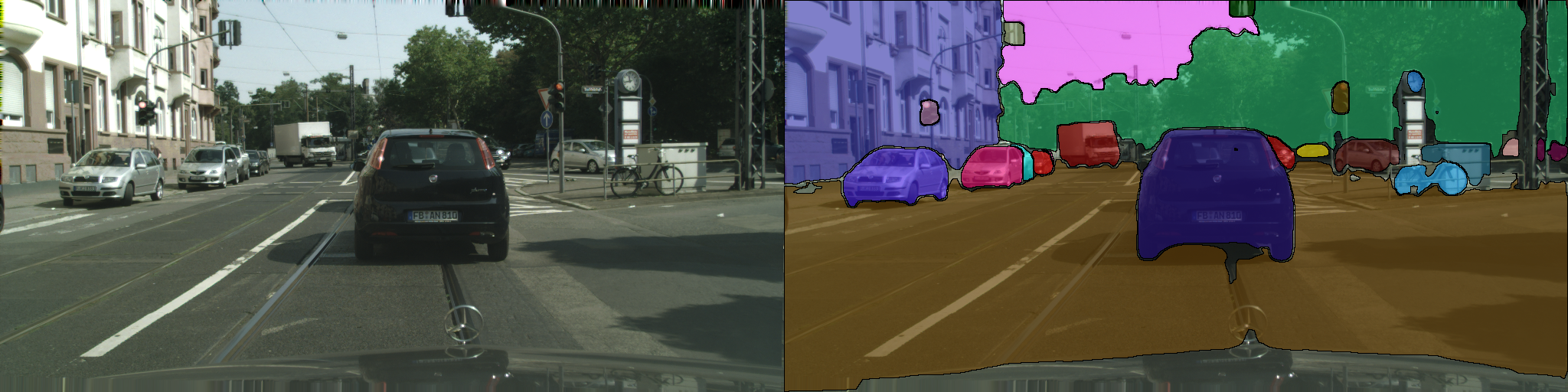}
\vspace{0.01in}
\includegraphics[width=0.95\linewidth]{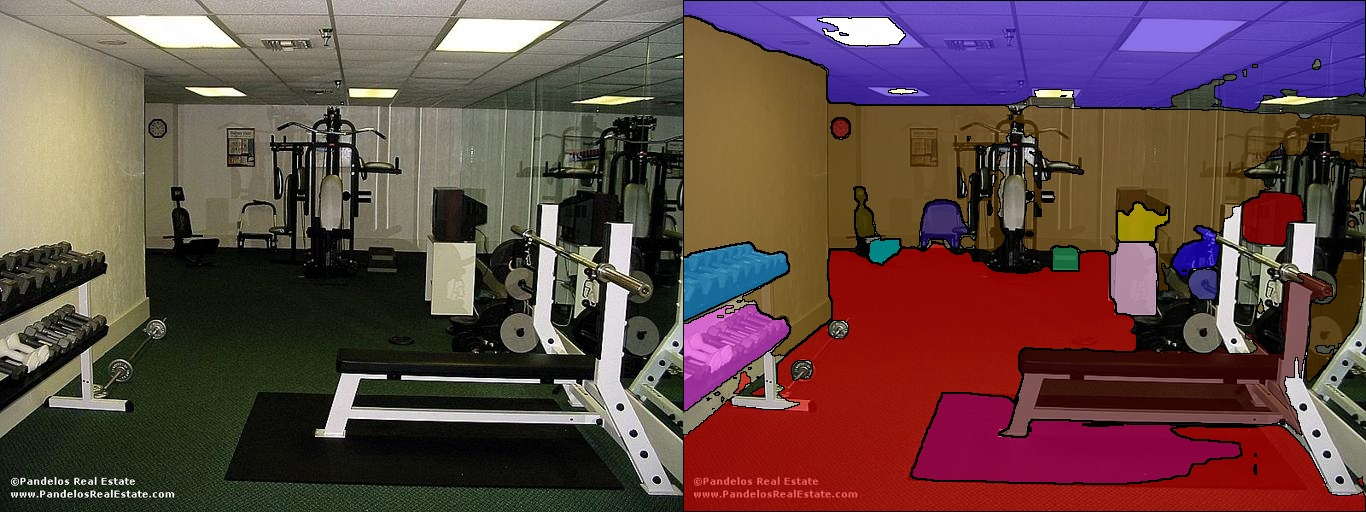}
\vspace{0.01in}
\includegraphics[width=0.95\linewidth]{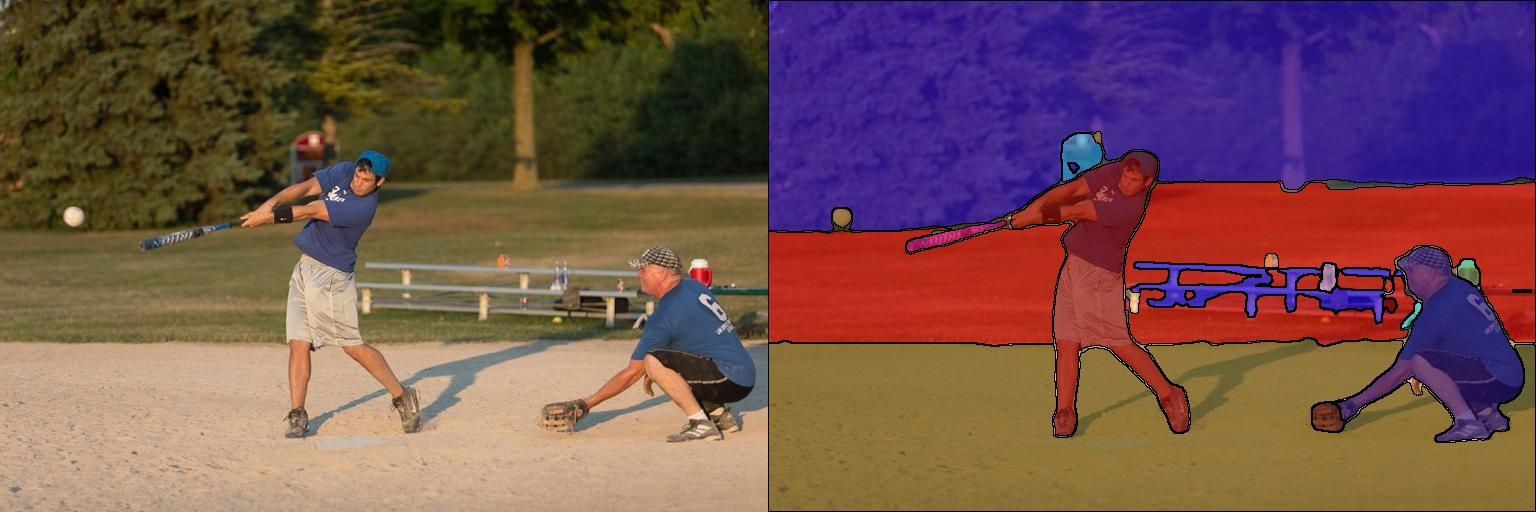}
\end{center}
\caption{The rows from up to bottom show the inference results of ADE20K, CityScapes, Places2 and Object365 with the same model trained in only COCO dataset.}
\label{vis_various_dataset}
\end{figure}

We also provide more visualization results of our entity segmentation approach on some detection and segmentation dataset ADE20K~\cite{zhou2017scene}, CityScapes~\cite{cordts2016cityscapes}, Places2~\cite{zhou2017places} and Object365~\cite{shao2019objects365} in Figure~\ref{vis_various_dataset}. Note that the model is only trained in once COCO dataset without additional fine-tuning on other datasets.

We note that the model in Figure~\ref{fig:generalization} and \ref{vis_various_dataset} are trained from scratch (with random init) on COCO, rather than ImageNet
 pretrained weights. When the model is trained with a long enough schedule, \textit{random init} is competitive to or better than ImageNet pretrained weights~\cite{he2019rethinking}. Figure 6's model is trained with a long schedule 72 epochs. Furthermore, using the model trained from scratch on COCO is more convincing for demonstrating our approach's excellent generalization benefits on ImageNet on data that has not been seen during training.

\section{Conclusion}
This paper proposes a new task named entity segmentation (ES) which is aimed to serve downstream applications such as image manipulation/editing that have high requirements for segmentation mask quality but not for class labels. In this regard, an entity is defined as any semantically-meaningful and -coherent segment in the image. In the absence of semantic labels, we design a new metric AP$_\text{e}$ to measure the ES performance. To effectively represent entities, we introduce a segmentation framework based on the unified center-based representation to handle both {\em thing} and {\em stuff} in this task. The proposed global kernel bank and overlap suppression modules further improve the segmentation performance. Experiments show that the models trained for ES task demonstrate strong generalization strengths and they perform exceptionally well in the open-world (cross-dataset) setting.

\newpage
{
	\small
	\bibliographystyle{ieee}
	\bibliography{egbib}

\begin{thebibliography}{10}\itemsep=-1pt

\bibitem{alexe2010object}
B.~Alexe, T.~Deselaers, and V.~Ferrari.
\newblock What is an object?
\newblock In {\em CVPR}, 2010.

\bibitem{barnes2009patchmatch}
C.~Barnes, E.~Shechtman, A.~Finkelstein, and D.~B. Goldman.
\newblock Patchmatch: A randomized correspondence algorithm for structural
  image editing.
\newblock {\em TOG}, 2009.

\bibitem{bendale2015towards}
A.~Bendale and T.~Boult.
\newblock Towards open world recognition.
\newblock In {\em CVPR}, 2015.

\bibitem{borji2019salient}
A.~Borji, M.-M. Cheng, Q.~Hou, H.~Jiang, and J.~Li.
\newblock Salient object detection: A survey.
\newblock {\em Computational visual media}, 2019.

\bibitem{borji2015salient}
A.~Borji, M.-M. Cheng, H.~Jiang, and J.~Li.
\newblock Salient object detection: A benchmark.
\newblock {\em TIP}, 2015.

\bibitem{cao2018pose}
K.~Cao, Y.~Rong, C.~Li, X.~Tang, and C.~C. Loy.
\newblock Pose-robust face recognition via deep residual equivariant mapping.
\newblock In {\em CVPR}, 2018.

\bibitem{carion2020end}
N.~Carion, F.~Massa, G.~Synnaeve, N.~Usunier, A.~Kirillov, and S.~Zagoruyko.
\newblock End-to-end object detection with transformers.
\newblock In {\em ECCV}, 2020.

\bibitem{chen2017rethinking}
L.-C. Chen, G.~Papandreou, F.~Schroff, and H.~Adam.
\newblock Rethinking atrous convolution for semantic image segmentation.
\newblock {\em arXiv:1706.05587}, 2017.

\bibitem{chen2018encoder}
L.-C. Chen, Y.~Zhu, G.~Papandreou, F.~Schroff, and H.~Adam.
\newblock Encoder-decoder with atrous separable convolution for semantic image
  segmentation.
\newblock In {\em ECCV}, 2018.

\bibitem{chen2020bi}
X.~Chen, K.-Y. Lin, J.~Wang, W.~Wu, C.~Qian, H.~Li, and G.~Zeng.
\newblock Bi-directional cross-modality feature propagation with
  separation-and-aggregation gate for rgb-d semantic segmentation.
\newblock In {\em ECCV}, 2020.

\bibitem{cheng2021per}
B.~Cheng, A.~G. Schwing, and A.~Kirillov.
\newblock Per-pixel classification is not all you need for semantic
  segmentation.
\newblock In {\em NeurlPS}, 2021.

\bibitem{cordts2016cityscapes}
M.~Cordts, M.~Omran, S.~Ramos, T.~Rehfeld, M.~Enzweiler, R.~Benenson,
  U.~Franke, S.~Roth, and B.~Schiele.
\newblock The cityscapes dataset for semantic urban scene understanding.
\newblock In {\em CVPR}, 2016.

\bibitem{dai2016instance}
J.~Dai, K.~He, and J.~Sun.
\newblock Instance-aware semantic segmentation via multi-task network cascades.
\newblock In {\em CVPR}, 2016.

\bibitem{dai2016rfcn}
J.~Dai, Y.~Li, K.~He, and J.~Sun.
\newblock {R-FCN:} object detection via region-based fully convolutional
  networks.
\newblock In {\em NeurIPS}, 2016.

\bibitem{dai2017deformable}
J.~Dai, H.~Qi, Y.~Xiong, Y.~Li, G.~Zhang, H.~Hu, and Y.~Wei.
\newblock Deformable convolutional networks.
\newblock In {\em ICCV}, 2017.

\bibitem{darabi2012image}
S.~Darabi, E.~Shechtman, C.~Barnes, D.~B. Goldman, and P.~Sen.
\newblock Image melding: Combining inconsistent images using patch-based
  synthesis.
\newblock {\em TOG}, 2012.

\bibitem{infant}
I.~C. L.~U. Davis).
\newblock {Infant Categorization Development}.
\newblock
  \url{https://oakeslab.ucdavis.edu/infant-categorization-development.html}.

\bibitem{AR2020theoverlooked}
A.~R. {Dhamija}, M.~{Günther}, J.~{Ventura}, and T.~E. {Boult}.
\newblock The overlooked elephant of object detection: Open set.
\newblock In {\em WACV}, 2020.

\bibitem{dollar2015fast}
P.~Doll{\'a}r and C.~L. Zitnick.
\newblock Fast edge detection using structured forests.
\newblock In {\em PAMI}, 2015.

\bibitem{girshick2014rich}
R.~Girshick, J.~Donahue, T.~Darrell, and J.~Malik.
\newblock Rich feature hierarchies for accurate object detection and semantic
  segmentation.
\newblock In {\em CVPR}, 2014.

\bibitem{he2019rethinking}
K.~He, R.~Girshick, and P.~Doll{\'a}r.
\newblock Rethinking imagenet pre-training.
\newblock In {\em Proceedings of the IEEE/CVF International Conference on
  Computer Vision}, pages 4918--4927, 2019.

\bibitem{he2017mask}
K.~He, G.~Gkioxari, P.~Doll{\'a}r, and R.~Girshick.
\newblock Mask r-cnn.
\newblock In {\em ICCV}, 2017.

\bibitem{he2016res}
K.~He, X.~Zhang, S.~Ren, and J.~Sun.
\newblock Deep residual learning for image recognition.
\newblock In {\em CVPR}, 2016.

\bibitem{hu2018squeeze}
J.~Hu, L.~Shen, and G.~Sun.
\newblock Squeeze-and-excitation networks.
\newblock In {\em CVPR}, 2018.

\bibitem{hu2018learning}
R.~{Hu}, P.~{Dollár}, K.~{He}, T.~{Darrell}, and R.~{Girshick}.
\newblock Learning to segment every thing.
\newblock In {\em CVPR}, 2018.

\bibitem{huang2017densely}
G.~Huang, Z.~Liu, L.~Van Der~Maaten, and K.~Q. Weinberger.
\newblock Densely connected convolutional networks.
\newblock In {\em CVPR}, 2017.

\bibitem{huang2019ccnet}
Z.~Huang, X.~Wang, L.~Huang, C.~Huang, Y.~Wei, and W.~Liu.
\newblock Ccnet: Criss-cross attention for semantic segmentation.
\newblock In {\em ICCV}, 2019.

\bibitem{isola2017image}
P.~Isola, J.-Y. Zhu, T.~Zhou, and A.~A. Efros.
\newblock Image-to-image translation with conditional adversarial networks.
\newblock In {\em CVPR}, 2017.

\bibitem{jaiswal2021class}
A.~Jaiswal, Y.~Wu, P.~Natarajan, and P.~Natarajan.
\newblock Class-agnostic object detection.
\newblock In {\em WACV}, 2021.

\bibitem{jia2016dynamic}
X.~Jia, B.~De~Brabandere, T.~Tuytelaars, and L.~Van~Gool.
\newblock Dynamic filter networks.
\newblock In {\em NeurIPS}, 2016.

\bibitem{jiang2013salient}
H.~Jiang, J.~Wang, Z.~Yuan, Y.~Wu, N.~Zheng, and S.~Li.
\newblock Salient object detection: A discriminative regional feature
  integration approach.
\newblock In {\em CVPR}, 2013.

\bibitem{joseph2021open}
K.~J. Joseph, S.~Khan, F.~S. Khan, and V.~N. Balasubramanian.
\newblock Towards open world object detection.
\newblock In {\em CVPR}, 2021.

\bibitem{kirillov2019panopticfpn}
A.~Kirillov, R.~Girshick, K.~He, and P.~Doll{\'a}r.
\newblock Panoptic feature pyramid networks.
\newblock In {\em CVPR}, 2019.

\bibitem{kirillov2019panoptic}
A.~Kirillov, K.~He, R.~Girshick, C.~Rother, and P.~Doll{\'a}r.
\newblock Panoptic segmentation.
\newblock In {\em CVPR}, 2019.

\bibitem{krizhevsky2017imagenet}
A.~Krizhevsky, I.~Sutskever, and G.~E. Hinton.
\newblock Imagenet classification with deep convolutional neural networks.
\newblock In {\em NeurIPS}, 2012.

\bibitem{levin2004colorization}
A.~Levin, D.~Lischinski, and Y.~Weiss.
\newblock Colorization using optimization.
\newblock {\em ACM SIGGRAPH}, 2004.

\bibitem{li2016deep}
G.~Li and Y.~Yu.
\newblock Deep contrast learning for salient object detection.
\newblock In {\em CVPR}, 2016.

\bibitem{li2016fully}
Y.~Li, H.~Qi, J.~Dai, X.~Ji, and Y.~Wei.
\newblock Fully convolutional instance-aware semantic segmentation.
\newblock In {\em CVPR}, 2017.

\bibitem{li2021fully}
Y.~Li, H.~Zhao, X.~Qi, Y.~Chen, L.~Qi, L.~Wang, Z.~Li, J.~Sun, and J.~Jia.
\newblock Fully convolutional networks for panoptic segmentation with
  point-based supervision.
\newblock {\em arXiv preprint arXiv:2108.07682}, 2021.

\bibitem{li2020fully}
Y.~Li, H.~Zhao, X.~Qi, L.~Wang, Z.~Li, J.~Sun, and J.~Jia.
\newblock Fully convolutional networks for panoptic segmentation.
\newblock In {\em CVPR}, 2021.

\bibitem{lin2017feature}
T.-Y. Lin, P.~Doll{\'a}r, R.~Girshick, K.~He, B.~Hariharan, and S.~Belongie.
\newblock Feature pyramid networks for object detection.
\newblock In {\em CVPR}, 2017.

\bibitem{lin2017focal}
T.-Y. Lin, P.~Goyal, R.~Girshick, K.~He, and P.~Doll{\'a}r.
\newblock Focal loss for dense object detection.
\newblock In {\em ICCV}, 2017.

\bibitem{lin2014microsoft}
T.-Y. Lin, M.~Maire, S.~Belongie, J.~Hays, P.~Perona, D.~Ramanan,
  P.~Doll{\'a}r, and C.~L. Zitnick.
\newblock Microsoft coco: Common objects in context.
\newblock In {\em ECCV}, 2014.

\bibitem{liu2019auto}
C.~Liu, L.-C. Chen, F.~Schroff, H.~Adam, W.~Hua, A.~L. Yuille, and L.~Fei-Fei.
\newblock Auto-deeplab: Hierarchical neural architecture search for semantic
  image segmentation.
\newblock In {\em CVPR}, 2019.

\bibitem{liu2019end}
H.~Liu, C.~Peng, C.~Yu, J.~Wang, X.~Liu, G.~Yu, and W.~Jiang.
\newblock An end-to-end network for panoptic segmentation.
\newblock In {\em CVPR}, 2019.

\bibitem{liu2018intriguing}
R.~Liu, J.~Lehman, P.~Molino, F.~P. Such, E.~Frank, A.~Sergeev, and
  J.~Yosinski.
\newblock An intriguing failing of convolutional neural networks and the
  coordconv solution.
\newblock In {\em NeurIPS}, 2018.

\bibitem{liu2018path}
S.~Liu, L.~Qi, H.~Qin, J.~Shi, and J.~Jia.
\newblock Path aggregation network for instance segmentation.
\newblock In {\em CVPR}, 2018.

\bibitem{liu2021swin}
Z.~Liu, Y.~Lin, Y.~Cao, H.~Hu, Y.~Wei, Z.~Zhang, S.~Lin, and B.~Guo.
\newblock Swin transformer: Hierarchical vision transformer using shifted
  windows.
\newblock In {\em ICCV}, 2021.

\bibitem{liu2019large}
Z.~Liu, Z.~Miao, X.~Zhan, J.~Wang, B.~Gong, and S.~X. Yu.
\newblock Large-scale long-tailed recognition in an open world.
\newblock In {\em CVPR}, 2019.

\bibitem{long2015fully}
J.~Long, E.~Shelhamer, and T.~Darrell.
\newblock Fully convolutional networks for semantic segmentation.
\newblock In {\em CVPR}, 2015.

\bibitem{Marr1982Vision}
D.~Marr.
\newblock {\em Vision: a computational investigation into the human
  representation and processing of visual information}.
\newblock San Francisco: W.H. Freeman. Print., 1982.

\bibitem{milletari2016fully}
F.~Milletari, N.~Navab, S.~Ahmadi, and V-net.
\newblock Fully convolutional neural networks for volumetric medical image
  segmentation.
\newblock In {\em 3DV}, 2016.

\bibitem{morrison2018cartman}
D.~Morrison, A.~W. Tow, M.~McTaggart, R.~Smith, N.~Kelly-Boxall, S.~Wade-McCue,
  J.~Erskine, R.~Grinover, A.~Gurman, T.~Hunn, et~al.
\newblock Cartman: The low-cost cartesian manipulator that won the amazon
  robotics challenge.
\newblock In {\em ICRA}, 2018.

\bibitem{perez2003poisson}
P.~P{\'e}rez, M.~Gangnet, and A.~Blake.
\newblock Poisson image editing.
\newblock {\em ACM SIGGRAPH}, 2003.

\bibitem{DeepMask15}
P.~O. Pinheiro, R.~Collobert, and P.~Doll\'{a}r.
\newblock Learning to segment object candidates.
\newblock In {\em NeurIPS}, 2015.

\bibitem{qi2019amodal}
L.~Qi, L.~Jiang, S.~Liu, X.~Shen, and J.~Jia.
\newblock Amodal instance segmentation with kins dataset.
\newblock In {\em CVPR}, 2019.

\bibitem{qi2021multi}
L.~Qi, J.~Kuen, J.~Gu, Z.~Lin, Y.~Wang, Y.~Chen, Y.~Li, and J.~Jia.
\newblock Multi-scale aligned distillation for low-resolution detection.
\newblock In {\em CVPR}, 2021.

\bibitem{qi2018sequential}
L.~Qi, S.~Liu, J.~Shi, and J.~Jia.
\newblock Sequential context encoding for duplicate removal.
\newblock In {\em Advances in Neural Information Processing Systems}, 2018.

\bibitem{qi2020pointins}
L.~Qi, Y.~Wang, Y.~Chen, Y.~Chen, X.~Zhang, J.~Sun, and J.~Jia.
\newblock Pointins: Point-based instance segmentation.
\newblock {\em TPAMI}, 2021.

\bibitem{qin2019basnet}
X.~Qin, Z.~Zhang, C.~Huang, C.~Gao, M.~Dehghan, and M.~Jagersand.
\newblock Basnet: Boundary-aware salient object detection.
\newblock In {\em CVPR}, 2019.

\bibitem{rahman2018zero}
S.~Rahman, S.~Khan, and F.~Porikli.
\newblock Zero-shot object detection: Learning to simultaneously recognize and
  localize novel concepts.
\newblock In {\em ACCV}, 2018.

\bibitem{ren2015faster}
S.~Ren, K.~He, R.~B. Girshick, and J.~Sun.
\newblock Faster {R-CNN:} towards real-time object detection with region
  proposal networks.
\newblock In {\em NeurIPS}, 2015.

\bibitem{rong2019delving}
Y.~Rong, Z.~Liu, C.~Li, K.~Cao, and C.~C. Loy.
\newblock Delving deep into hybrid annotations for 3d human recovery in the
  wild.
\newblock In {\em ICCV}, 2019.

\bibitem{rong2020frankmocap}
Y.~Rong, T.~Shiratori, and H.~Joo.
\newblock Frankmocap: Fast monocular 3d hand and body motion capture by
  regression and integration.
\newblock {\em arXiv preprint arXiv:2008.08324}, 2020.

\bibitem{shao2019objects365}
S.~Shao, Z.~Li, T.~Zhang, C.~Peng, G.~Yu, X.~Zhang, J.~Li, and J.~Sun.
\newblock Objects365: A large-scale, high-quality dataset for object detection.
\newblock In {\em Proceedings of the IEEE/CVF International Conference on
  Computer Vision}, pages 8430--8439, 2019.

\bibitem{shu2014human}
G.~Shu.
\newblock Human detection, tracking and segmentation in surveillance video.
\newblock 2014.

\bibitem{simonyan2014very}
K.~Simonyan and A.~Zisserman.
\newblock Very deep convolutional networks for large-scale image recognition.
\newblock In {\em ICLR}, 2015.

\bibitem{sun2005image}
J.~Sun, L.~Yuan, J.~Jia, and H.-Y. Shum.
\newblock Image completion with structure propagation.
\newblock {\em TOG}, 2005.

\bibitem{szegedy2016inception}
C.~Szegedy, S.~Ioffe, V.~Vanhoucke, and A.~Alemi.
\newblock Inception-v4, inception-resnet and the impact of residual connections
  on learning.
\newblock In {\em arXiv}, 2016.

\bibitem{szegedy2016rethinking}
C.~Szegedy, V.~Vanhoucke, S.~Ioffe, J.~Shlens, and Z.~Wojna.
\newblock Rethinking the inception architecture for computer vision.
\newblock In {\em CVPR}, 2016.

\bibitem{szeliski2010computer}
R.~Szeliski.
\newblock {\em Computer vision: algorithms and applications}.
\newblock Springer Science \& Business Media, 2010.

\bibitem{tian2020conditional}
Z.~Tian, C.~Shen, and H.~Chen.
\newblock Conditional convolutions for instance segmentation.
\newblock In {\em ECCV}, 2020.

\bibitem{tian2019fcos}
Z.~Tian, C.~Shen, H.~Chen, and T.~He.
\newblock Fcos: Fully convolutional one-stage object detection.
\newblock In {\em ICCV}, 2019.

\bibitem{ullah2020brief}
I.~Ullah, M.~Jian, S.~Hussain, J.~Guo, H.~Yu, X.~Wang, and Y.~Yin.
\newblock A brief survey of visual saliency detection.
\newblock {\em Multimedia Tools and Applications}, 2020.

\bibitem{wang2017high}
T.-C. Wang, M.-Y. Liu, J.-Y. Zhu, A.~Tao, J.~Kautz, and B.~Catanzaro.
\newblock High-resolution image synthesis and semantic manipulation with
  conditional gans.
\newblock {\em arXiv:1711.11585}, 2017.

\bibitem{wang2021unidentified}
W.~Wang, M.~Feiszli, H.~Wang, and D.~Tran.
\newblock Unidentified video objects: A benchmark for dense, open-world
  segmentation.
\newblock {\em arXiv preprint arXiv:2104.04691}, 2021.

\bibitem{wang2019solo}
X.~Wang, T.~Kong, C.~Shen, Y.~Jiang, and L.~Li.
\newblock Solo: Segmenting objects by locations.
\newblock In {\em ECCV}, 2020.

\bibitem{wang2020solov2}
X.~Wang, R.~Zhang, T.~Kong, L.~Li, and C.~Shen.
\newblock Solov2: Dynamic and fast instance segmentation.
\newblock In {\em NeurIPS}, 2020.

\bibitem{wang2018inpainting}
Y.~Wang, X.~Tao, X.~Qi, X.~Shen, and J.~Jia.
\newblock Image inpainting via generative multi-column convolutional neural
  networks.
\newblock In {\em NeurIPS}, 2018.

\bibitem{xie2020polarmask}
E.~Xie, P.~Sun, X.~Song, W.~Wang, X.~Liu, D.~Liang, C.~Shen, and P.~Luo.
\newblock Polarmask: Single shot instance segmentation with polar
  representation.
\newblock In {\em CVPR}, 2020.

\bibitem{xie2021segformer}
E.~Xie, W.~Wang, Z.~Yu, A.~Anandkumar, J.~M. Alvarez, and P.~Luo.
\newblock Segformer: Simple and efficient design for semantic segmentation with
  transformers.
\newblock In {\em NeurIPS}, 2022.

\bibitem{xing2020malleable}
Y.~Xing, J.~Wang, and G.~Zeng.
\newblock Malleable 2.5 d convolution: Learning receptive fields along the
  depth-axis for rgb-d scene parsing.
\newblock In {\em ECCV}, 2020.

\bibitem{xiong2019upsnet}
Y.~Xiong, R.~Liao, H.~Zhao, R.~Hu, M.~Bai, E.~Yumer, and R.~Urtasun.
\newblock Upsnet: A unified panoptic segmentation network.
\newblock In {\em CVPR}, 2019.

\bibitem{yu2020context}
C.~Yu, J.~Wang, C.~Gao, G.~Yu, C.~Shen, and N.~Sang.
\newblock Context prior for scene segmentation.
\newblock In {\em CVPR}, 2020.

\bibitem{yu2018bisenet}
C.~Yu, J.~Wang, C.~Peng, C.~Gao, G.~Yu, and N.~Sang.
\newblock Bisenet: Bilateral segmentation network for real-time semantic
  segmentation.
\newblock In {\em ECCV}, 2018.

\bibitem{yu2018learning}
C.~Yu, J.~Wang, C.~Peng, C.~Gao, G.~Yu, and N.~Sang.
\newblock Learning a discriminative feature network for semantic segmentation.
\newblock In {\em CVPR}, 2018.

\bibitem{yu2018generative}
J.~Yu, Z.~Lin, J.~Yang, X.~Shen, X.~Lu, and T.~S. Huang.
\newblock Generative image inpainting with contextual attention.
\newblock In {\em CVPR}, 2018.

\bibitem{yu2019free}
J.~Yu, Z.~Lin, J.~Yang, X.~Shen, X.~Lu, and T.~S. Huang.
\newblock Free-form image inpainting with gated convolution.
\newblock In {\em ICCV}, 2019.

\bibitem{zhang2021k}
W.~Zhang, J.~Pang, K.~Chen, and C.~C. Loy.
\newblock K-net: Towards unified image segmentation.
\newblock {\em NeurlPS}, 2021.

\bibitem{zhao2017pyramid}
H.~Zhao, J.~Shi, X.~Qi, X.~Wang, and J.~Jia.
\newblock Pyramid scene parsing network.
\newblock In {\em CVPR}, 2017.

\bibitem{zhao2018psanet}
H.~Zhao, Y.~Zhang, S.~Liu, J.~Shi, C.~Change~Loy, D.~Lin, and J.~Jia.
\newblock Psanet: Point-wise spatial attention network for scene parsing.
\newblock In {\em ECCV}, 2018.

\bibitem{zhou2017places}
B.~Zhou, A.~Lapedriza, A.~Khosla, A.~Oliva, and A.~Torralba.
\newblock Places: A 10 million image database for scene recognition.
\newblock {\em IEEE transactions on pattern analysis and machine intelligence},
  40(6):1452--1464, 2017.

\bibitem{zhou2017scene}
B.~Zhou, H.~Zhao, X.~Puig, S.~Fidler, A.~Barriuso, and A.~Torralba.
\newblock Scene parsing through ade20k dataset.
\newblock In {\em CVPR}, 2017.

\bibitem{zhu2019deformable}
X.~Zhu, H.~Hu, S.~Lin, and J.~Dai.
\newblock Deformable convnets v2: More deformable, better results.
\newblock In {\em CVPR}, 2019.

\end{thebibliography}
}

\begin{IEEEbiography}	[{\includegraphics[width=1in,height=1.25in,clip,keepaspectratio]
		{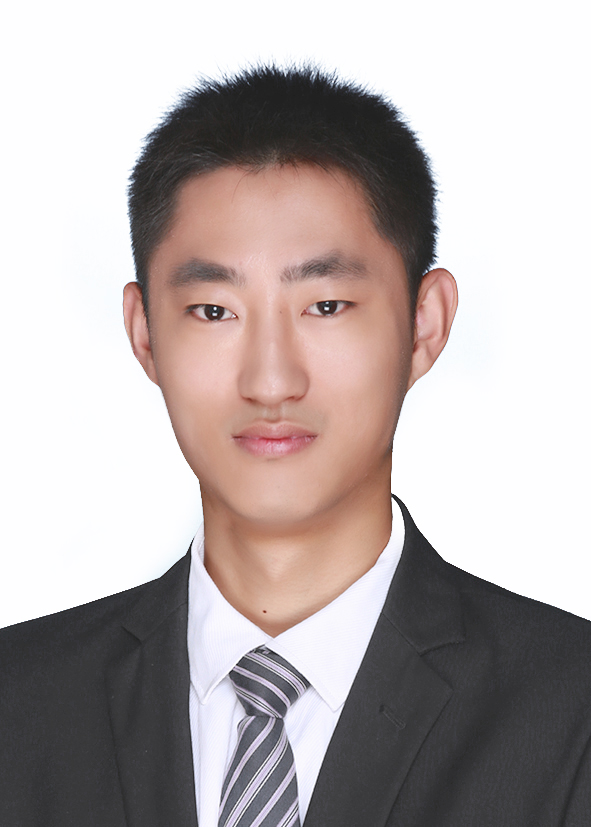}}]
	{Lu QI} received the Ph.D. degree in The Chinese University of Hong Kong in 2021. He is currently working as Postdoc with Prof. Ming-Hsuan Yang at UC Merced. He served as senior program chair of AAAI2023. His current research interests include instance-level detection, image generation and cross-modal training.
\end{IEEEbiography}

\begin{IEEEbiography}
	[{\includegraphics[width=1in,height=1.5in,clip,keepaspectratio]
	    {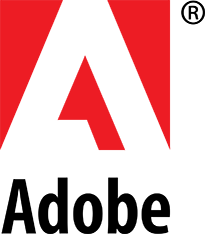}}]
	{Jason Kuen} is a computer vision and deep learning researcher in the Creative Intelligence Lab (CIL) at Adobe Research. He completed his PhD at Nanyang Technological University (NTU), Singapore in 2019 and obtained his bachelor’s degree from Multimedia University (MMU), Malaysia in 2014. He is interested in (very) large-scale visual recognition problems as well as deep learning techniques for computer vision under limited data and/or computational resource settings. 
\end{IEEEbiography}

\begin{IEEEbiography}
	[{\includegraphics[width=1in,height=1.5in,clip,keepaspectratio]
	    {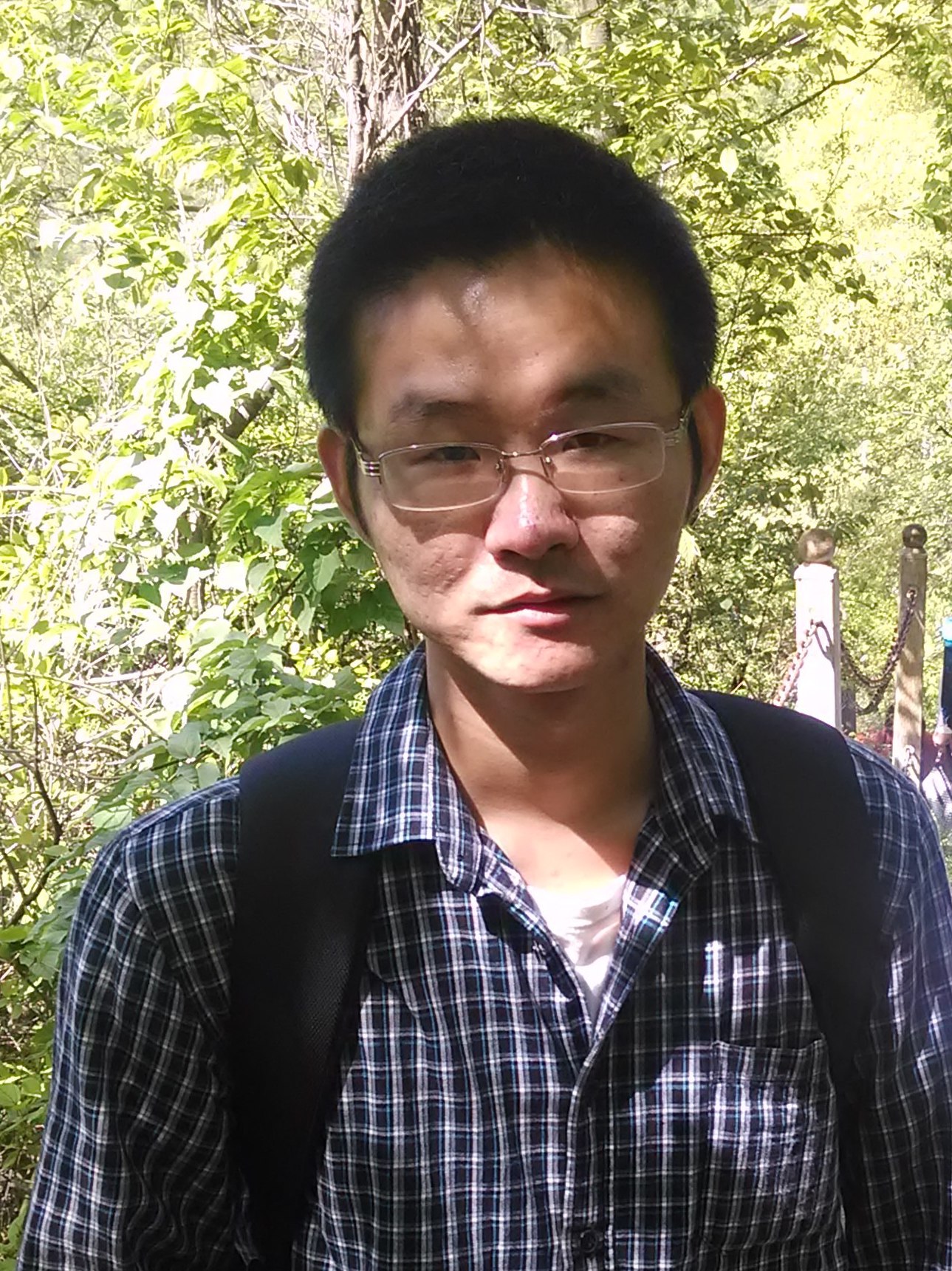}}]
	{Yi Wang} is a research scientist in Shanghai AI Lab. He received the Ph.D. degree at The Chinese University of Hong Kong in 2021. He is currently working as a research scientist in Shanghai AI Lab. His current research interests include video understanding and generation models.
\end{IEEEbiography}

\begin{IEEEbiography}
	[{\includegraphics[width=1in,height=1.5in,clip,keepaspectratio]
	    {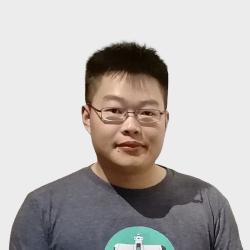}}]
	{Jiuxiang Gu} is a Research Scientist in the Document Intelligence Lab of Adobe Research. He obtained my Ph.D. degree in 2019 at Nanyang Technological University, Singapore, under the supervision of Prof. Jiangfei Cai, Dr. Gang Wang and Prof. Tsuhan Chen. His recent reserach topic includes: Document Understanding and Reasoning, Uncertainty and Out-of-Distribution, Self-Supervised Learning, Human-in-the-loop and Privacy Preserving.
\end{IEEEbiography}

\begin{IEEEbiography}
	[{\includegraphics[width=1in,height=1.5in,clip,keepaspectratio]
	    {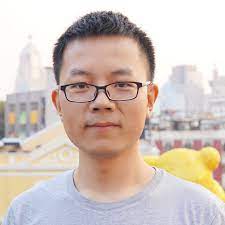}}]
	{Hengshuang Zhao} is currently a postdoctoral researcher at Computer Science and Artificial Intelligence Laboratory (CSAIL) at MIT, working with Prof. Antonio Torralba. Before that, he was a postdoctoral researcher at Torr Vision Group in the Department of Engineering Science at the University of Oxford, working with Prof. Philip Torr. He obtained my Ph.D. degree in the Department of Computer Science and Engineering at The Chinese University of Hong Kong, supervised by Prof. Jiaya Jia. His general research interests cover the broad area of computer vision and machine learning.
\end{IEEEbiography}

\begin{IEEEbiography}
	[{\includegraphics[width=1in,height=1.5in,clip,keepaspectratio]
	    {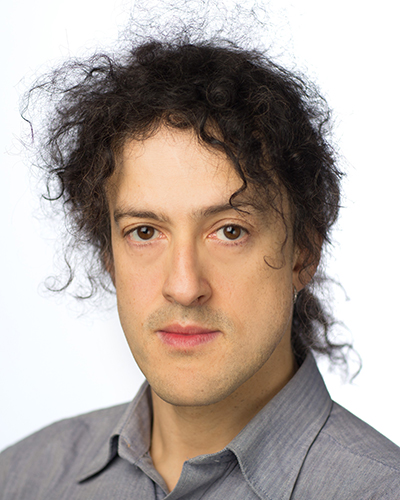}}]
	{Philip Torr}
	received the PhD degree from
Oxford University. After working for another
three years at Oxford, he worked for six years
for Microsoft Research, first in Redmond, then
in Cambridge, founding the vision side of
the Machine Learning and Perception Group.
He is now a professor at Oxford University.
He has won awards from top vision conferences, including ICCV, CVPR, ECCV, NIPS and BMVC. He is a senior member of the
IEEE and a Royal Society Wolfson Research
Merit Award holder.
\end{IEEEbiography}

\begin{IEEEbiography}
	[{\includegraphics[width=1in,height=1.5in,clip,keepaspectratio]
	    {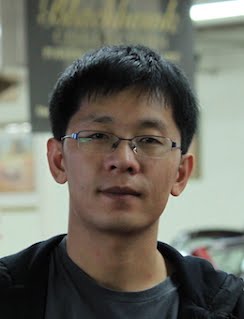}}]
	{Zhe Lin} is a Senior Principal Research Scientist in Imagination Lab, Adobe Research. He received his Ph.D. degree in electrical engineering from University of Maryland at College Park in May 2009. Prior to that, he obtained his M.S. degree in electrical engineering from Korea Advanced Institute of Science and Technology in August 2004, and B.Eng. degree in automatic control from University of Science and Technology of China. He served as area chairs for ICCV, CVPR, ECCV, and several other conferences for organization. His research interests include computer vision, image and video processing, machine learning, deep learning, artificial intelligence. 
\end{IEEEbiography}

\begin{IEEEbiography}
	[{\includegraphics[width=1in,height=1.25in,clip,keepaspectratio]
		{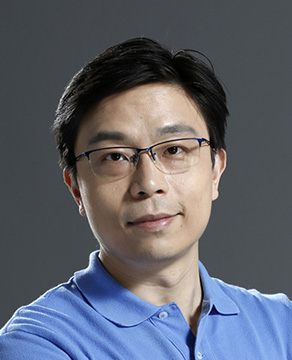}}]
	{Jiaya Jia}
	received the Ph.D. degree in Computer Science from Hong Kong University of Science and Technology in 2004 and is currently a full professor in Department of Computer Science and Engineering at the Chinese University of Hong Kong (CUHK). He is in the editorial boards of IEEE Transactions on Pattern Analysis and Machine Intelligence (TPAMI) and International Journal of Computer Vision (IJCV). He continuously served as area chairs for ICCV, CVPR, AAAI, ECCV, and several other conferences for organization. He was on program committees of major conferences in graphics and computational imaging, including ICCP, SIGGRAPH, and SIGGRAPH Asia. He is a Fellow of the IEEE. 
\end{IEEEbiography}
\end{document}